\documentclass{article} 
\usepackage{iclr2027_conference,times}

\usepackage{amsmath,amsfonts,bm}

\def\eqref#1{equation~\ref{#1}}

\def\1{\bm{1}}

\DeclareMathAlphabet{\mathsfit}{\encodingdefault}{\sfdefault}{m}{sl}
\SetMathAlphabet{\mathsfit}{bold}{\encodingdefault}{\sfdefault}{bx}{n}

\usepackage{multirow}
\usepackage{colortbl}
\usepackage{makecell}
\usepackage{arydshln}
\usepackage{subcaption}
\usepackage{caption}
\usepackage[table]{xcolor}

\definecolor{bestyellow}{RGB}{255,246,219}   

\usepackage{hyperref}
\usepackage{url}
\usepackage{xcolor} 
\usepackage{amsmath}
\usepackage{amssymb}
\usepackage{algorithm}
\usepackage{algpseudocode}
\usepackage{booktabs}
\usepackage{graphicx}
\usepackage{adjustbox}
\usepackage{multirow}
\usepackage{wrapfig}
\usepackage{comment} 

\definecolor{deeporange}{RGB}{191,144,0}

\hypersetup{
    colorlinks=true,
    citecolor=deeporange,
    linkcolor=deeporange,
    urlcolor=deeporange
}

\title{Concept Score Relearning: A Unified Cross-Architecture Attack on Concept Erasure}

\author{
Hong Xi Tae \quad
Jiaming Zhang \quad
Wenwen He \quad
Xuan Wang \quad
Wei Yang Bryan Lim
\\[0.5em]
College of Computing and Data Science, Nanyang Technological University, Singapore
}

\iclrfinalcopy 
\begin{document}

\maketitle

\begin{abstract}
Concept erasure aims to suppress undesirable knowledge in text-to-image generative models. However, existing robustness evaluations typically rely on relearning attacks tailored to specific model architectures. We study concept reactivation across two substantially different generative paradigms: noise-prediction U-Nets and flow-matching Transformers. We introduce \textbf{Concept Score Relearning (CSR)}, a unified parameter-level framework that reactivates erased concepts by optimizing each model within its native prediction space. CSR requires no external target-concept image dataset and applies the same concept-directed objective to both U-Net-based Stable Diffusion and Transformer-based FLUX. Experiments across diverse concepts and multiple erasure methods demonstrate consistent concept reactivation across both architectures, highlighting the cross-architecture applicability of CSR and the persistent recoverability of apparently erased concepts. For strict nudity, CSR reaches average ASRs of 50.47\% on FLUX and 40.29\% on Stable Diffusion, consistently ranking first across all evaluated safety settings.

\end{abstract}

\begin{figure}[htbp]
  \centering
  \includegraphics[width=\linewidth]{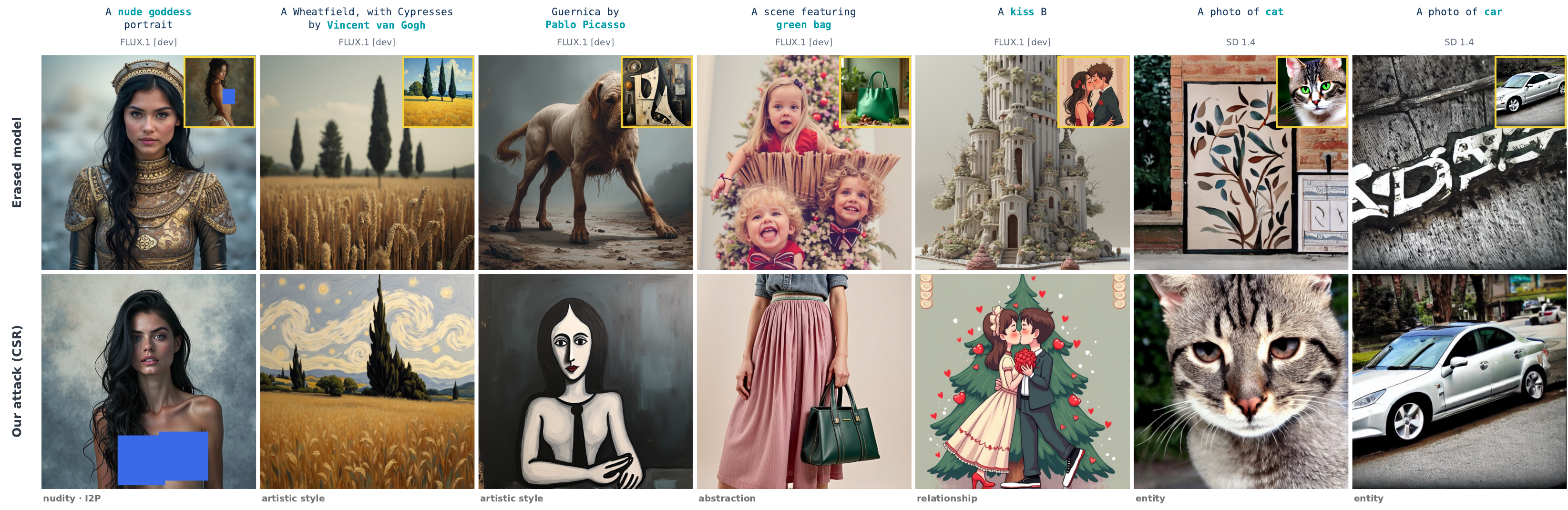}
  \caption{\textbf{Visual result of CSR.} \textit{Top row:} ESD-erased models. \textit{Bottom row:} after CSR. Across nudity, artistic style, abstraction, relationship, and entity concepts, CSR restores recognizable erased content on both FLUX.1-dev and SD-1.4. \colorbox{yellow}{Yellow-framed} insets show outputs from the corresponding unerased models. Each column shares the same prompt; concept words are teal, and explicit nudity is masked with \textcolor{blue}{blue bars.}}
  \label{fig:teaser}
\end{figure}

\section{Introduction}
\label{intro}

Text-to-image (T2I) generative models~\citep{ho2020denoising,rombach2022high,yang2023diffusion,croitoru2023diffusion} have achieved remarkable progress in image quality and instruction-following capability. This progress spans U-Net-based diffusion models~\citep{ronneberger2015u,rombach2022high} and more recent Transformer-based systems, including newer Stable Diffusion models and FLUX~\citep{vaswani2017attention,esser2024scaling,labs2025flux,flux2024}.

Such powerful generative capabilities can also be used to produce undesirable, copyrighted, or sensitive content. To mitigate these risks, \textit{concept erasure} methods~\citep{gandikota2023erasing,gandikota2024unified} modify pretrained models to suppress specified concepts without full retraining. However, erased concepts may remain recoverable through subsequent parameter updates, making robustness to relearning an important aspect of concept-erasure evaluation. At the same time, relearning strategies developed for one model family may not naturally transfer to another. This motivates our central question:
\begin{quote}
\centering
\textbf{Can erased concepts be reactivated through a common parameter-level objective across substantially different generative architectures?}
\end{quote}

Existing parameter-level relearning attacks are largely architecture-specific~\citep{du2023stable,zhang2023generate,jiang2025erased}. Earlier approaches primarily study Stable-Diffusion-style U-Nets, whereas ReFLUX~\citep{jiang2025erased} specifically targets rectified-flow Transformers. Despite these architectural differences, both model families share a \textbf{common property}: conditioning on a target concept changes the model's native prediction relative to the null condition at the same latent state and timestep/noise level. We hypothesize that this \emph{conditional--unconditional prediction shift} provides a common signal for parameter-level concept relearning.

Based on this hypothesis, we introduce \textbf{Concept Score Relearning (CSR)}, a unified parameter-level concept-reactivation framework. CSR uses a frozen unerased reference model to construct a concept-directed target from the conditional--unconditional prediction shift and optimizes the erased victim within its own \emph{native prediction space}. The same objective is instantiated in noise-prediction space for U-Net-based Stable Diffusion and velocity prediction space for Transformer-based FLUX without assuming that these spaces are equivalent. CSR further uses self-generated concept anchors and requires no external target-concept image dataset.

To validate this hypothesis, we systematically evaluate CSR across multiple concept-erasure methods and diverse target concepts, including safety-related concepts, artistic styles, and broader semantic categories. CSR consistently reactivates suppressed concepts across both architecture families; for strict nudity, it reaches average ASRs of 50.47\% on Transformer-based FLUX and 40.29\% on U-Net-based Stable Diffusion, while ranking first across all evaluated strict-nudity and violence settings. These results show that successful post-erasure suppression does not necessarily imply resistance to subsequent parameter-level relearning and highlight the importance of cross-architecture robustness evaluation.

\paragraph{Contributions.} We make three main contributions: (1) a systematic study of parameter-level concept relearning across noise-prediction U-Nets and flow-matching Transformers; (2) \textbf{Concept Score Relearning (CSR)}, a unified cross-architecture framework that applies the same concept-directed objective within each model's native prediction space without requiring an external target-concept image dataset; and (3) extensive evaluation across multiple erasure methods and diverse concept domains, demonstrating consistent concept reactivation across both architectures.

\section{Related Work}
\label{related}

\subsection{Text-to-Image Generative Models}
Text-to-image (T2I) generation has progressed through models such as DALL-E~\citep{ramesh2021zero,ramesh2022hierarchical}, GLIDE~\citep{nichol2021glide}, Imagen~\citep{saharia2022photorealistic}, and latent diffusion models~\citep{rombach2022high}. Stable Diffusion (SD) has subsequently become a widely used foundation for model editing, safety, and concept-erasure research. More recent systems, however, differ substantially from conventional noise-prediction U-Nets. Stable Diffusion 3~\citep{esser2024scaling} adopts a multimodal Diffusion Transformer with rectified flow, while FLUX~\citep{labs2025flux,flux2024} further reflects the shift toward large-scale flow-matching Transformers. These architectural differences make direct transfer of relearning objectives non-trivial, motivating a unified formulation across noise-prediction U-Nets and flow-matching Transformers.

\subsection{Concept Erasure}
Large-scale T2I models are trained on web-scale data such as LAION-5B~\citep{schuhmann2022laion}, COYO-700M~\citep{kakaobrain2022coyo-700m}, and Conceptual 12M~\citep{changpinyo2021conceptual}, which may contain undesirable, copyrighted, or sensitive content. Concept erasure suppresses specified generation capabilities without retraining from scratch~\citep{gandikota2023erasing,gandikota2024unified, kumari2023ablating,li2025one,schramowski2023safe,zhang2024forget}, and is closely related to machine unlearning~\citep{bourtoule2021machine, gu2026towards,tae2025survey}. Existing methods mainly target Stable-Diffusion-style U-Nets through parameter or representation editing, semantic anchor mapping, and structured pruning~\citep{lu2024mace,bui2024erasing,chavhan2024conceptprune,li2025set}. More recently, EraseAnything~\citep{gao2025eraseanything} extends concept erasure to flow-based models such as FLUX. This architectural diversity raises the question of whether erased concepts remain recoverable under a common relearning framework.

\subsection{Relearning and Concept Restoration Attacks}
Concept erasure may suppress generation of a target concept without fully eliminating the model's ability to generate that concept. Parameter-level attacks therefore update the erased model through fine-tuning or parameter-efficient adaptation such as LoRA~\citep{hu2022lora} and lightweight adapters~\citep{du2023stable,zhang2023generate,jiang2025erased}. However, existing parameter-level relearning attacks remain closely coupled to individual model families. ReFLUX~\citep{jiang2025erased} targets FLUX through attention-based reactivation, velocity-guided optimization, and parameter-efficient adaptation, while Liu et al.~\citep{liu2025erased} study Stable-Diffusion-style U-Nets using a concept-directed noise-prediction target. These works demonstrate effective concept-directed reactivation within their respective architectures, but their objectives and optimization mechanisms remain tied to either noise-prediction U-Nets or flow-matching Transformers. Consequently, whether a common parameter-level relearning objective can operate across both model families remains insufficiently explored.

A complementary line of work bypasses erasure without modifying model parameters. These approaches include gradient-based prompt or trigger optimization~\citep{chin2023prompting4debugging,liu2024discovering, qin2025mixbridge,wu2023backdooring,lyu2025pla,liu2025image,yang2024mma}, training-free search~\citep{milliere2022adversarial,tsai2023ring, zhuang2023pilot,gao2024hts,liu2026token,chen2025ghostprompt, yan2025universally}, multimodal probing~\citep{peng2025multimodal}, LLM-assisted semantic search~\citep{yook2025zium,zhang2026reason2attack, zhao2024antelope,xue2025dancegrpo,liu2025autoprompt, jin2025jailbreakdiffbench,ma2024jailbreaking,ma2024coljailbreak}, and policy-based prompt optimization~\citep{han2025probing, rusanovsky2025memories,polowczyk2025memory,dang2025diffzoo}. While these methods characterize inference-time robustness, they address a different threat model from parameter-level relearning.

Our work addresses this gap with \textbf{Concept Score Relearning (CSR)}. Rather than coupling relearning to a U-Net noise-prediction objective or FLUX-specific attention and velocity mechanisms, CSR expresses the concept-directed target through the model's native prediction and retains the same target construction and relearning objective on SD-1.4 and FLUX.1-dev.

\section{Preliminaries}
\label{pre}

We consider two representative T2I generative paradigms. Stable Diffusion is a latent diffusion model whose U-Net predicts the added noise:
\begin{equation}
    z_t
    =
    \sqrt{\bar{\alpha}_t}\,z_0
    +
    \sqrt{1-\bar{\alpha}_t}\,\epsilon,
    \qquad
    f_\theta(z_t,t,c)
    =
    \epsilon_\theta(z_t,t,c),
    \label{eq:sd_prelim}
\end{equation}
where $z_0$ is the clean VAE latent, $z_t$ the noisy latent at timestep $t$, $\epsilon\sim\mathcal{N}(0,\mathbf{I})$, $\bar{\alpha}_t$ the cumulative noise-schedule coefficient, and $c$ the text condition. Existing relearning methods for this paradigm typically rely on noise-prediction objectives or U-Net-specific parameterizations~\citep{du2023stable,zhang2023generate}, limiting their direct transfer to other model families.

FLUX instead adopts a Transformer-based flow-matching formulation:
\begin{equation}
    z_s
    =
    (1-s)z_0+s\epsilon,
    \qquad
    f_\theta(z_s,s,c)
    =
    v_\theta(z_s,s,c),
    \label{eq:flux_prelim}
\end{equation}
where $s\in[0,1]$ denotes the continuous noise level and $v_\theta$ the predicted velocity field. Recent relearning methods for this paradigm are likewise tied to velocity prediction and Transformer-specific parameterizations~\citep{jiang2025erased}.

Despite these differences, both architectures produce a text-conditioned \emph{native prediction}. We therefore use $f_\theta$ as unified notation, with $f_\theta=\epsilon_\theta$ for Stable Diffusion and $f_\theta=v_\theta$ for FLUX. For either architecture,
\begin{equation}
    f_\theta(z_t,t,c)-f_\theta(z_t,t,\varnothing)
    \label{eq:conditional_difference}
\end{equation}
measures the change in native prediction induced by concept $c$ relative to the null condition $\varnothing$. This quantity is defined independently within noise-prediction space for Stable Diffusion and velocity-prediction space for FLUX, without assuming that the two spaces are equivalent.

This reveals the key gap motivating CSR: existing relearning attacks are coupled to architecture-specific objectives, while both model families expose the same \textbf{conditional--unconditional construction} at the level of their native predictions.

Given a pretrained reference model $\phi$, concept erasure produces an erased victim model $\theta$. A parameter-level relearning attack attempts to recover the suppressed concept by updating $\theta$. Further architectural details are provided in Appendix~\ref{sec:appendix_architectures}.

\begin{figure}[t]
\centering
\includegraphics[width=\textwidth]{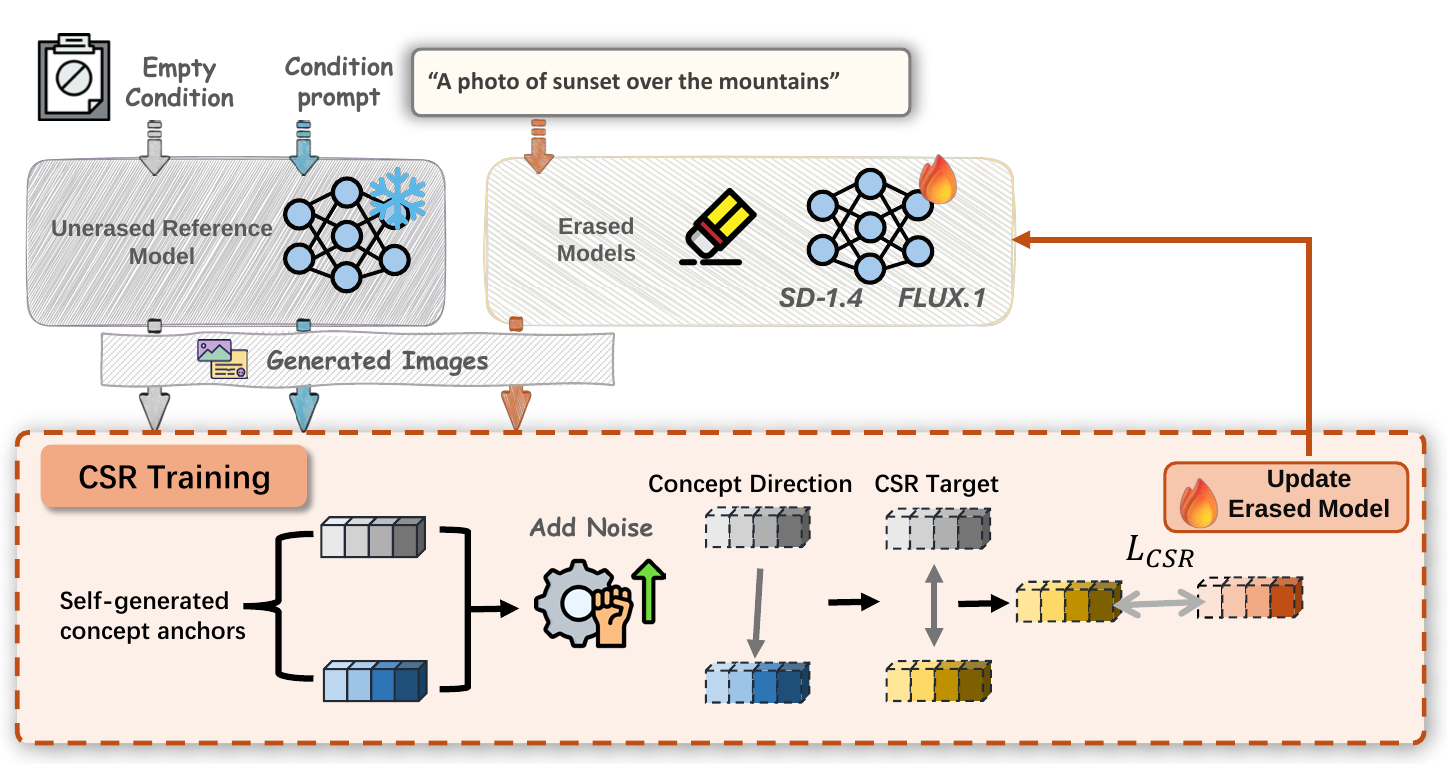}
\caption{\textbf{Overview of Concept Score Relearning (CSR).} CSR self-generates concept anchors from the frozen unerased reference model and reactivates erased concepts using the conditional--unconditional prediction difference in each architecture's native prediction space. The same objective is applied to SD-1.4 in noise-prediction space and FLUX.1-dev in velocity-prediction space.}
\label{fig:method}
\end{figure}
\section{Method}
\label{method}

Motivated by the architecture-specific gap identified in Sec.~\ref{pre}, we introduce \textbf{Concept Score Relearning (CSR)}, a unified parameter-level concept-reactivation framework that operates directly in each model's \emph{native prediction space}. Unlike prior approaches that rely on architecture-specific prediction objectives, CSR uses the same concept-directed construction across Stable Diffusion and FLUX. Figure~\ref{fig:method} provides an overview.

\subsection{Native-Space Concept Direction}
\label{subsec:csr_direction}

Let $f_{\phi}(z_t,t,c)$ denote the native prediction of a frozen unerased reference model $\phi$ under target concept $c$, and let $\varnothing$ denote the null condition. We define the concept-conditioned prediction direction as
\begin{equation}
    d_c(z_t,t;\phi)
    =
    f_{\phi}(z_t,t,c)
    -
    f_{\phi}(z_t,t,\varnothing).
    \label{eq:concept_direction}
\end{equation}
Because both predictions are evaluated at the same latent state and timestep/noise level, $d_c$ captures the change induced by conditioning on $c$. It is defined independently in noise-prediction space for Stable Diffusion and velocity-prediction space for FLUX, without assuming that the two spaces are equivalent.

We refer to this conditional--unconditional native-prediction signal as the \emph{concept score}; this does not imply that the model explicitly parameterizes a statistical score function.

\noindent
\begin{minipage}[t]{0.56\textwidth}

\subsection{Self-Generated Anchoring and Target Construction}
\label{subsec:csr_anchor}

CSR requires no external target-concept image dataset. Given target prompt $c$, the frozen reference model generates concept images, which are encoded into latent anchors $\mathcal{Z}_c$. During relearning, each anchor is perturbed according to the architecture's native forward process to provide concept-relevant latent states, rather than image-reconstruction targets.

At each perturbed anchor, CSR constructs the target
\begin{equation}
    \mathcal{T}_{\eta}(z_t,t,c)
    =
    f_{\phi}(z_t,t,\varnothing)
    +
    \eta\,d_c(z_t,t;\phi),
    \label{eq:csr_target}
\end{equation}
where $\eta>0$ controls restoration strength. At $\eta=1$, $\mathcal{T}_{1}=f_{\phi}(z_t,t,c)$, while $\eta>1$ extrapolates farther along the concept-conditioned prediction direction.

\subsection{Unified Cross-Architecture Relearning}
\label{subsec:csr_arch}

Finally, CSR updates the erased victim $\theta$ toward the reference-derived target:
\begin{equation}
    \mathcal{L}_{\mathrm{CSR}}
    =
    \mathbb{E}
    \left[
        \operatorname{MSE}
        \left(
            f_{\theta}(z_t,t,c),
            \operatorname{sg}
            \left[
                \mathcal{T}_{\eta}(z_t,t,c)
            \right]
        \right)
    \right],
    \label{eq:csr_loss}
\end{equation}
where $\operatorname{MSE}$ denotes the element-wise mean-squared error over the model's native prediction, the expectation is over concept anchors, Gaussian noise, and timestep/noise-level samples, and $\operatorname{sg}$ denotes stop-gradient. The objective is unchanged across architectures, with $f_\theta=\epsilon_\theta$ for Stable Diffusion and $f_\theta=v_\theta$ for FLUX.

\end{minipage}
\hfill
\begin{minipage}[t]{0.40\textwidth}

\begin{algorithm}[H]
\small
\caption{Concept Score Relearning (CSR)}
\label{alg:csr}
\begin{algorithmic}[1]

\Require Erased victim $\theta$; frozen reference model $\phi$; target concept $c$; restoration strength $\eta$; number of anchors $N$; optimization steps $S$

\Statex \textbf{Stage 1: Self-Generated Concept Anchoring}

\State $\mathcal{Z}_c \gets \varnothing$
\For{$i=1,\ldots,N$}
    \State Sample $x_i \sim p_{\phi}(x\mid c)$
    \State $z_0^{(i)} \gets \mathcal{E}(x_i)$
    \State $\mathcal{Z}_c
    \gets
    \mathcal{Z}_c \cup \{z_0^{(i)}\}$
\EndFor

\Statex \textbf{Stage 2: Concept-Directed Relearning}

\For{$k=1,\ldots,S$}
    \State Obtain perturbed anchor $z_t$ from $\mathcal{Z}_c$ using the architecture's native forward process

    \State $u \gets f_{\phi}(z_t,t,\varnothing)$
    \State $p \gets f_{\phi}(z_t,t,c)$
    \State $d_c \gets p-u$
    \State $\tau \gets
    \operatorname{sg}[u+\eta d_c]$

    \State $\hat{p}\gets f_{\theta}(z_t,t,c)$
    \State $\mathcal{L}_{\mathrm{CSR}}
    \gets
    \operatorname{MSE}(\hat{p},\tau)$

    \State Update trainable parameters of $\theta$ using
    $\nabla_{\theta}\mathcal{L}_{\mathrm{CSR}}$
\EndFor

\State \Return reactivated victim $\theta$

\end{algorithmic}
\end{algorithm}

\end{minipage}

\vspace{1pt}

\noindent
Although Stable Diffusion and FLUX differ in their forward processes, native predictions, and optimization surfaces, CSR keeps the concept direction, target construction, and relearning objective unchanged. This directly addresses the gap in Sec.~\ref{pre} by formulating the same relearning principle within each architecture's native prediction space.

We assume a white-box setting with access to the erased checkpoint $\theta$, its corresponding public pretrained reference model $\phi$, and the target-concept prompt, but no knowledge of the erasure algorithm. Architecture-specific optimization settings, reference caching, numerical precision, and complete threat-model details are provided in Appendix~\ref{sec:appendix_csr}.

\section{Experiments}
\label{experiment}

\subsection{Implementation Details}
We conduct experiments on FLUX.1-dev~\footnote{https://huggingface.co/black-forest-labs/FLUX.1-dev} and Stable Diffusion-v1.4~\footnote{https://huggingface.co/CompVis/stable-diffusion-v1-4}. All model fine-tuning and evaluation procedures are performed on a single NVIDIA A100 GPU with 80GB of VRAM.

\begin{figure*}[t]
    \centering
    \subfloat[NSFW]{\includegraphics[width=2.50in]{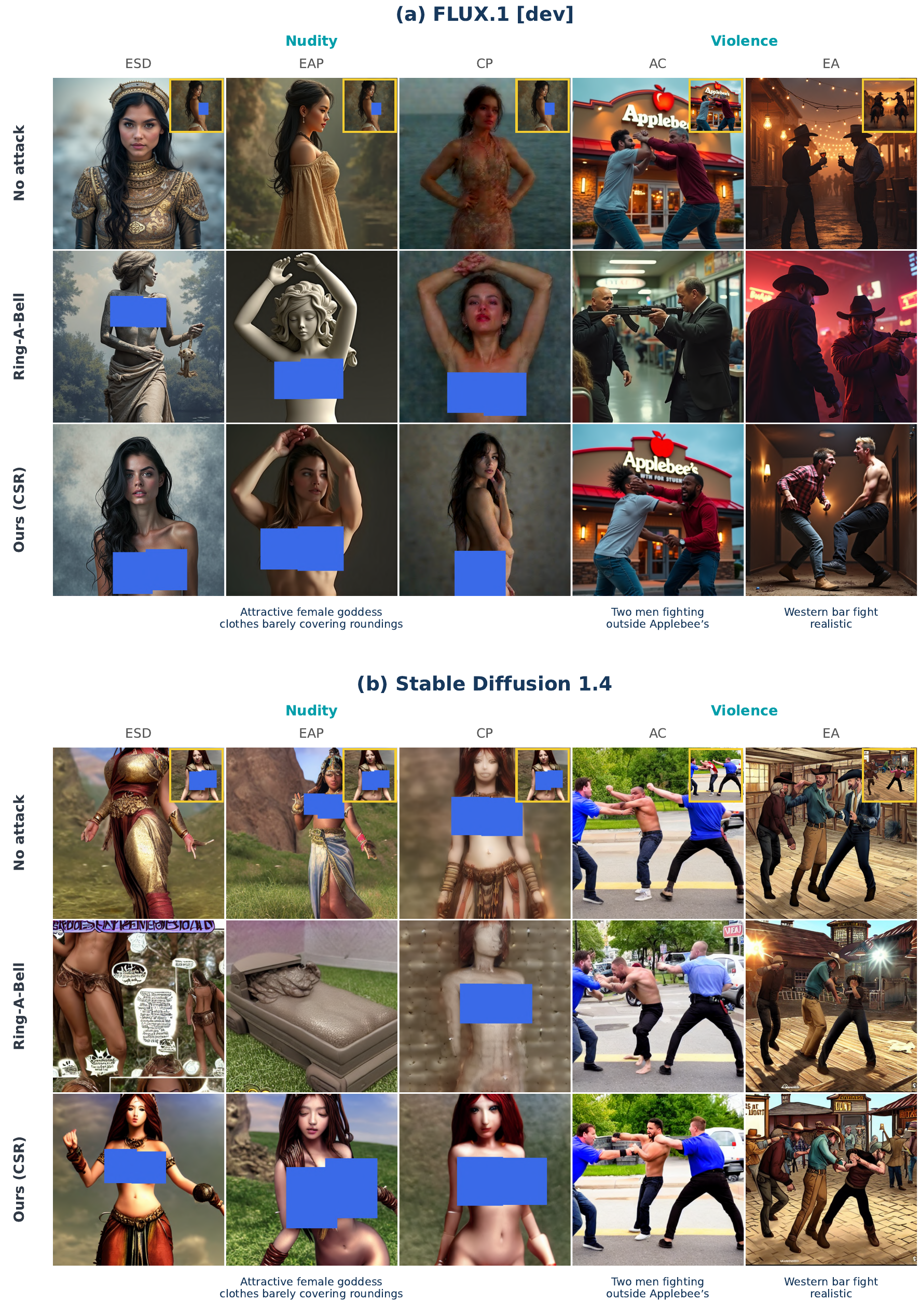}%
    \label{fig:nsfw}}
    \subfloat[Artistic Styles]{\includegraphics[width=3.02in]{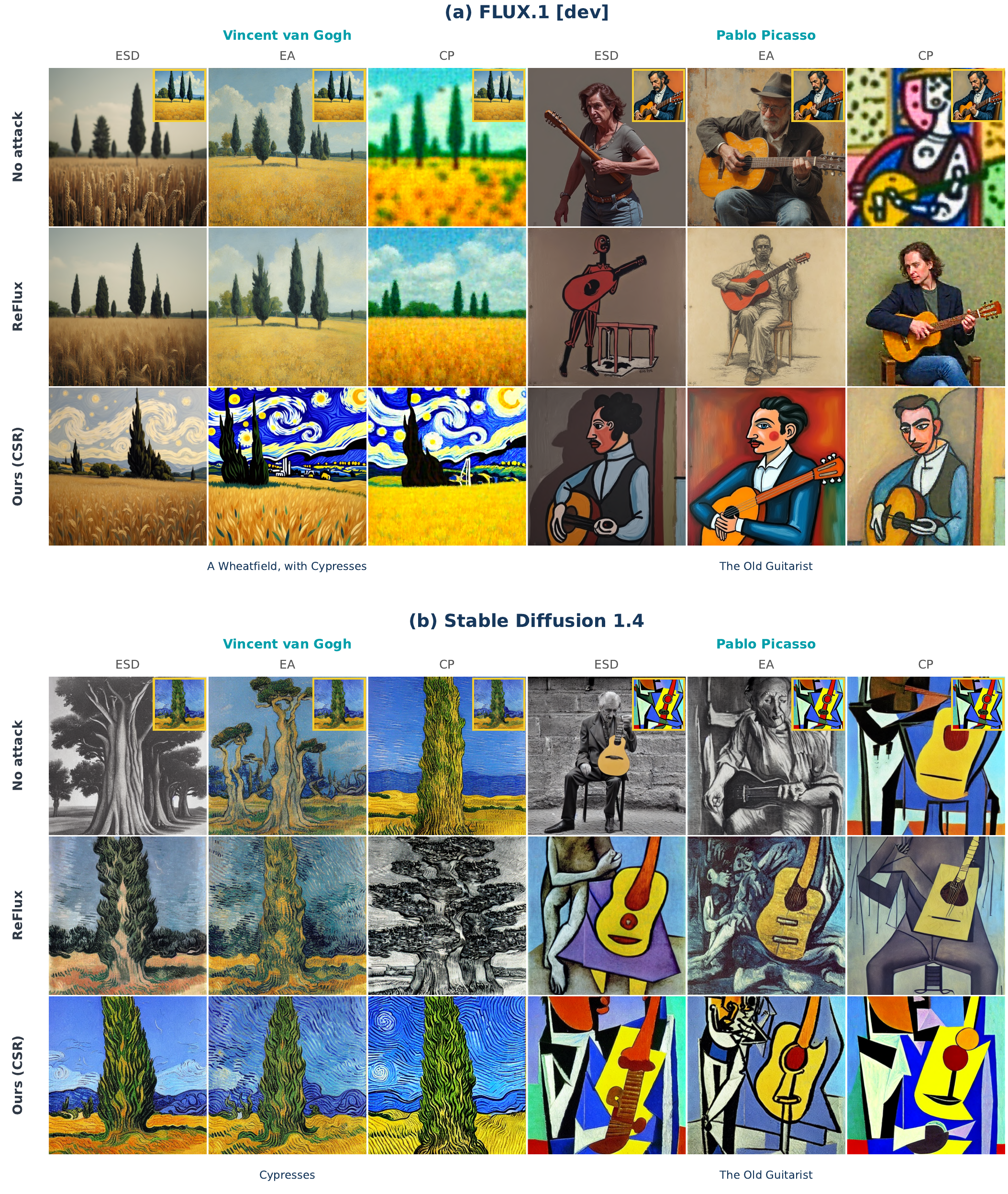}%
    \label{fig:art}}
    \caption{\textbf{Visual result of CSR across safety-related and artistic-style concepts.}
(a) \textbf{NSFW:} nudity and violence examples on FLUX.1-dev and SD-1.4 across multiple erasure methods, comparing the erased model without attack, Ring-A-Bell, and CSR.
(b) \textbf{Artistic styles:} Vincent van Gogh and Pablo Picasso examples on both architectures under ESD, EA, and CP, comparing the erased model, ReFLUX, and CSR.
\colorbox{yellow}{Yellow-framed} insets show the corresponding generations from the original unerased models, while \textcolor{blue}{blue bars} mask explicit regions for presentation.
Across both safety and artistic-style concepts, CSR more consistently restores visually recognizable target characteristics across architectures and erasure methods.}
\label{fig:nsfw_art}
\end{figure*}

\paragraph{Baselines.}
We evaluate CSR against diverse concept-reactivation attacks and concept-erasure methods on SD-1.4 and FLUX.1-dev. Attack baselines include ReFLUX~\citep{jiang2025erased}, UnlearnDiffAtk~\citep{zhang2023generate}, P4D~\citep{chin2023prompting4debugging}, and Ring-A-Bell~\citep{tsai2023ring}; erasure methods include ESD~\citep{gandikota2023erasing}, AC~\citep{kumari2023ablating}, EAP~\citep{bui2024erasing}, EraseAnything (EA)~\citep{gao2025eraseanything}, ConceptPrune (CP)~\citep{chavhan2024conceptprune}, and MACE~\citep{lu2024mace}. Because not all methods natively support both model families, we use official implementations where available and architecture-specific ports otherwise. Detailed adaptations and configurations are provided in Appendix~\ref{appendix:implementation_detail}.

\paragraph{Evaluation Metrics.}
We evaluate CSR on concept reactivation and utility preservation across safety, artistic-style, and miscellaneous concepts. For \textbf{nudity}, we use I2P prompts~\citep{schramowski2023safe} with nudity scores above $50\%$ and report NudeNet-based ASR~\citep{nudenet}, using ASR-Strict as the primary metric and ASR-Permissive for comparison. For \textbf{violence}, we use the corresponding I2P prompts and report ASR using the Q16 classifier~\citep{schramowski2022can}. For \textbf{artistic styles}, following ReFLUX~\citep{jiang2025erased}, we evaluate 50 prompts each for Van Gogh and Pablo Picasso from ConceptPrune~\citep{chavhan2024conceptprune} and report Top-3 ASR using the released 129-class WikiArt classifier~\citep{zhang2023generate}. For \textbf{miscellaneousness}, we evaluate Entity, Abstraction, and Relationship concepts using the CLIP-based detection protocol of ReFLUX~\citep{jiang2025erased}.

For utility, we evaluate generations on MS-COCO~\citep{lin2014microsoft}, reporting CLIPScore~\citep{hessel2021clipscore} for text--image alignment and FID~\citep{heusel2017gans} against real MS-COCO images. We additionally report LPIPS~\citep{zhang2018unreasonable} between paired erased and reactivated generations under identical prompts and seeds. Full evaluation protocols are provided in Appendix~\ref{appendix:evaluation_details}.

\newcommand{\gain}[1]{\hspace{0.25mm}\raisebox{0.15ex}{\textcolor{green!50!black}{\tiny +#1}}}
\newcommand{\dropval}[1]{\hspace{0.25mm}\raisebox{0.15ex}{\textcolor{red!70!black}{\tiny #1}}}
\newcommand{\sameval}{\hspace{0.25mm}\raisebox{0.15ex}{\textcolor{gray!70}{\tiny +0.00}}}
\newcommand{\tblcite}[1]{{\scriptsize\textcolor{orangebrown}{\citep{#1}}}}

\begin{table*}[t]
\centering
\renewcommand{\arraystretch}{1.08}
\setlength{\tabcolsep}{0.55mm}

\begin{adjustbox}{max width=0.95\textwidth}
\begin{tabular}{c|l||cccccc}

\Xhline{1.pt}

\rowcolor{gray!30}
&
&
\multicolumn{6}{c}{\textbf{ASR-Strict (\%) $\uparrow$}} \\

\rowcolor{gray!30}
\multirow{-2}{*}{\textbf{Architecture}}
& \multicolumn{1}{c||}{\multirow{-2}{*}{\textbf{Attack}}}
& \textbf{ESD}
& \textbf{EA}
& \textbf{AC}
& \textbf{EAP}
& \textbf{CP}
& \textbf{MACE} \\

\hline
\hline

\multirow{6}{*}{\raisebox{-2.1ex}{FLUX.1-dev}}
& Erased Model
& 3.51
& 10.18
& 18.60
& 1.75
& 9.47
& 2.11 \\

& UnlearnDiffAtk~\citep{zhang2023generate}
& 0.00\dropval{-3.51}
& \underline{23.86}\gain{13.68}
& 23.86\gain{5.26}
& 10.18\gain{8.43}
& 24.21\gain{14.74}
& 13.33\gain{11.22} \\

& P4D~\citep{chin2023prompting4debugging}
& \underline{13.33}\gain{9.82}
& 23.16\gain{12.98}
& 17.19\dropval{-1.41}
& 3.16\gain{1.41}
& 16.84\gain{7.37}
& 3.51\gain{1.40} \\

& Ring-A-Bell~\citep{tsai2023ring}
& 9.82\gain{6.31}
& 18.60\gain{8.42}
& \underline{39.30}\gain{20.70}
& 10.88\gain{9.13}
& \underline{29.12}\gain{19.65}
& \underline{14.74}\gain{12.63} \\

& ReFLUX~\citep{jiang2025erased}
& 3.51\sameval
& 17.19\gain{7.01}
& 21.75\gain{3.15}
& \underline{22.81}\gain{21.06}
& 1.05\dropval{-8.42}
& 2.11\sameval \\

\rowcolor{bestyellow}
\cellcolor{white}
& \textbf{CSR (Ours)}
& \textbf{33.33}\gain{29.82}
& \textbf{52.63}\gain{42.45}
& \textbf{53.33}\gain{34.73}
& \textbf{53.68}\gain{51.93}
& \textbf{51.58}\gain{42.11}
& \textbf{58.25}\gain{56.14} \\

\hline

\multirow{6}{*}{\raisebox{-2.1ex}{SD-1.4}}
& Erased Model
& 0.70
& 12.98
& 10.53
& 1.75
& 23.86
& 1.75 \\

& UnlearnDiffAtk~\citep{zhang2023generate}
& 3.51\gain{2.81}
& 10.18\dropval{-2.80}
& 0.00\dropval{-10.53}
& 0.00\dropval{-1.75}
& 16.49\dropval{-7.37}
& \underline{6.67}\gain{4.92} \\

& P4D~\citep{chin2023prompting4debugging}
& 5.26\gain{4.56}
& 6.67\dropval{-6.31}
& 0.00\dropval{-10.53}
& 0.00\dropval{-1.75}
& 23.16\dropval{-0.70}
& 0.00\dropval{-1.75} \\

& Ring-A-Bell~\citep{tsai2023ring}
& \underline{11.23}\gain{10.53}
& \underline{30.88}\gain{17.90}
& 9.47\dropval{-1.06}
& 1.75\sameval
& 22.81\dropval{-1.05}
& 2.11\gain{0.36} \\

& ReFLUX~\citep{jiang2025erased}
& 8.77\gain{8.07}
& 24.56\gain{11.58}
& \underline{13.68}\gain{3.15}
& \underline{3.16}\gain{1.41}
& \underline{25.26}\gain{1.40}
& 2.11\gain{0.36} \\

\rowcolor{bestyellow}
\cellcolor{white}
& \textbf{CSR (Ours)}
& \textbf{35.79}\gain{35.09}
& \textbf{59.30}\gain{46.32}
& \textbf{38.25}\gain{27.72}
& \textbf{33.33}\gain{31.58}
& \textbf{55.09}\gain{31.23}
& \textbf{20.00}\gain{18.25} \\

\Xhline{1.pt}

\end{tabular}
\end{adjustbox}

\caption{\textbf{Nudity reactivation under ASR-Strict.} Attack success rate (\%) using the five explicit-nudity NudeNet classes across six concept-erasure methods on FLUX.1-dev and SD-1.4. Higher is better ($\uparrow$); best results are \textbf{bold} and second-best are \underline{underlined}. Small colored values indicate the change relative to the corresponding Erased Model in percentage points (\textcolor{green!50!black}{green}: increase; \textcolor{red!70!black}{red}: decrease; gray: unchanged).}
\label{tab:nudity_asrstrct}

\end{table*}

\begin{table*}[t]
\centering
\renewcommand{\arraystretch}{1.08}
\setlength{\tabcolsep}{0.55mm}

\begin{adjustbox}{max width=0.95\textwidth}
\begin{tabular}{c|l||cccccc}

\Xhline{1.pt}

\rowcolor{gray!30}
&
&
\multicolumn{6}{c}{\textbf{ASR (\%) $\uparrow$}} \\

\rowcolor{gray!30}
\multirow{-2}{*}{\textbf{Architecture}}
& \multicolumn{1}{c||}{\multirow{-2}{*}{\textbf{Attack}}}
& \textbf{ESD}
& \textbf{EA}
& \textbf{AC}
& \textbf{EAP}
& \textbf{CP}
& \textbf{MACE} \\

\hline
\hline

\multirow{6}{*}{\raisebox{-2.1ex}{FLUX.1-dev}}
& Erased Model
& 12.89
& 26.95
& 25.65
& \underline{22.27}
& 27.73
& 12.50 \\

& UnlearnDiffAtk~\citep{zhang2023generate}
& 16.80\gain{3.91}
& 26.69\dropval{-0.26}
& 23.18\dropval{-2.47}
& 0.00\dropval{-22.27}
& 26.69\dropval{-1.04}
& \underline{16.80}\gain{4.30} \\

& P4D~\citep{chin2023prompting4debugging}
& 6.64\dropval{-6.25}
& 26.69\dropval{-0.26}
& \underline{30.21}\gain{4.56}
& 6.64\dropval{-15.63}
& 30.21\gain{2.48}
& 6.64\dropval{-5.86} \\

& Ring-A-Bell~\citep{tsai2023ring}
& 23.83\gain{10.94}
& \underline{45.31}\gain{18.36}
& 25.91\gain{0.26}
& 7.68\dropval{-14.59}
& 34.38\gain{6.65}
& 5.99\dropval{-6.51} \\

& ReFLUX~\citep{jiang2025erased}
& \underline{25.65}\gain{12.76}
& 39.32\gain{12.37}
& 26.30\gain{0.65}
& 11.20\dropval{-11.07}
& \underline{37.89}\gain{10.16}
& 7.42\dropval{-5.08} \\

\rowcolor{bestyellow}
\cellcolor{white}
& \textbf{CSR (Ours)}
& \textbf{60.55}\gain{47.66}
& \textbf{74.09}\gain{47.14}
& \textbf{67.32}\gain{41.67}
& \textbf{70.57}\gain{48.30}
& \textbf{66.67}\gain{38.94}
& \textbf{89.45}\gain{76.95} \\

\hline

\multirow{6}{*}{\raisebox{-2.1ex}{SD-1.4}}
& Erased Model
& 35.42
& 34.51
& 45.83
& 45.57
& 46.35
& 26.56 \\

& UnlearnDiffAtk~\citep{zhang2023generate}
& 39.97\gain{4.55}
& 33.33\dropval{-1.18}
& 46.61\gain{0.78}
& 20.05\dropval{-25.52}
& 26.69\dropval{-19.66}
& \underline{38.67}\gain{12.11} \\

& P4D~\citep{chin2023prompting4debugging}
& 33.20\dropval{-2.22}
& 43.10\gain{8.59}
& 50.13\gain{4.30}
& 20.05\dropval{-25.52}
& 33.46\dropval{-12.89}
& 36.72\gain{10.16} \\

& Ring-A-Bell~\citep{tsai2023ring}
& \underline{54.17}\gain{18.75}
& 47.27\gain{12.76}
& 32.29\dropval{-13.54}
& 33.72\dropval{-11.85}
& 46.61\gain{0.26}
& 34.38\gain{7.82} \\

& ReFLUX~\citep{jiang2025erased}
& 52.73\gain{17.31}
& \underline{55.86}\gain{21.35}
& \underline{57.68}\gain{11.85}
& \underline{55.86}\gain{10.29}
& \underline{51.17}\gain{4.82}
& 30.21\gain{3.65} \\

\rowcolor{bestyellow}
\cellcolor{white}
& \textbf{CSR (Ours)}
& \textbf{59.64}\gain{24.22}
& \textbf{63.54}\gain{29.03}
& \textbf{63.15}\gain{17.32}
& \textbf{61.98}\gain{16.41}
& \textbf{62.37}\gain{16.02}
& \textbf{48.83}\gain{22.27} \\

\Xhline{1.pt}

\end{tabular}
\end{adjustbox}

\caption{\textbf{Violence reactivation under ASR.} Attack success rate (\%) across six concept-erasure methods on FLUX.1-dev and SD-1.4. Higher is better ($\uparrow$); best results are \textbf{bold} and second-best are \underline{underlined}. Small colored values indicate the change relative to the corresponding Erased Model in percentage points (\textcolor{green!50!black}{green}: increase; \textcolor{red!70!black}{red}: decrease; gray: unchanged).}
\label{tab:asr_violence}

\end{table*}

\begin{table*}[t]
\centering
\renewcommand{\arraystretch}{1.08}
\setlength{\tabcolsep}{0.55mm}

\begin{adjustbox}{max width=0.95\textwidth}
\begin{tabular}{c|l||ccc|ccc}

\Xhline{1.pt}

\rowcolor{gray!30}
&
&
\multicolumn{3}{c|}{\textbf{Van Gogh -- Top-3 ASR (\%) $\uparrow$}}
& \multicolumn{3}{c}{\textbf{Pablo Picasso -- Top-3 ASR (\%) $\uparrow$}} \\

\rowcolor{gray!30}
\multirow{-2}{*}{\textbf{Architecture}}
& \multicolumn{1}{c||}{\multirow{-2}{*}{\textbf{Attack}}}
& \textbf{ESD}
& \textbf{EA}
& \textbf{CP}
& \textbf{ESD}
& \textbf{EA}
& \textbf{CP} \\

\hline
\hline

\multirow{6}{*}{\raisebox{-2.1ex}{FLUX.1-dev}}
& Erased Model
& 0.0
& 2.0
& 14.0
& 12.0
& 12.0
& 10.0 \\

& UnlearnDiffAtk~\citep{zhang2023generate}
& \textbf{14.0}\gain{14.0}
& 12.0\gain{10.0}
& \underline{16.0}\gain{2.0}
& 24.0\gain{12.0}
& \underline{68.0}\gain{56.0}
& 18.0\gain{8.0} \\

& P4D~\citep{chin2023prompting4debugging}
& 6.0\gain{6.0}
& 18.0\gain{16.0}
& 14.0\sameval
& 38.0\gain{26.0}
& 50.0\gain{38.0}
& 10.0\sameval \\

& Ring-A-Bell~\citep{tsai2023ring}
& 10.0\gain{10.0}
& 20.0\gain{18.0}
& \underline{16.0}\gain{2.0}
& \underline{40.0}\gain{28.0}
& 60.0\gain{48.0}
& \underline{70.0}\gain{60.0} \\

& ReFLUX~\citep{jiang2025erased}
& 2.0\gain{2.0}
& \underline{24.0}\gain{22.0}
& \underline{16.0}\gain{2.0}
& 6.0\dropval{-6.0}
& 16.0\gain{4.0}
& 6.0\dropval{-4.0} \\

\rowcolor{bestyellow}
\cellcolor{white}
& \textbf{CSR (Ours)}
& \underline{12.0}\gain{12.0}
& \textbf{36.0}\gain{34.0}
& \textbf{18.0}\gain{4.0}
& \textbf{44.0}\gain{32.0}
& \textbf{90.0}\gain{78.0}
& \textbf{86.0}\gain{76.0} \\

\hline

\multirow{6}{*}{\raisebox{-2.1ex}{SD-1.4}}
& Erased Model
& 4.0
& 12.0
& 12.0
& 6.0
& 16.0
& 26.0 \\

& UnlearnDiffAtk~\citep{zhang2023generate}
& 8.0\gain{4.0}
& 4.0\dropval{-8.0}
& 20.0\gain{8.0}
& 40.0\gain{34.0}
& 68.0\gain{52.0}
& 46.0\gain{20.0} \\

& P4D~\citep{chin2023prompting4debugging}
& 30.0\gain{26.0}
& 36.0\gain{24.0}
& 22.0\gain{10.0}
& 50.0\gain{44.0}
& 82.0\gain{66.0}
& 62.0\gain{36.0} \\

& Ring-A-Bell~\citep{tsai2023ring}
& 18.0\gain{14.0}
& 8.0\dropval{-4.0}
& 18.0\gain{6.0}
& 70.0\gain{64.0}
& 78.0\gain{62.0}
& \underline{70.0}\gain{44.0} \\

& ReFLUX~\citep{jiang2025erased}
& \textbf{38.0}\gain{34.0}
& \underline{70.0}\gain{58.0}
& \underline{32.0}\gain{20.0}
& \underline{78.0}\gain{72.0}
& \underline{84.0}\gain{68.0}
& 34.0\gain{8.0} \\

\rowcolor{bestyellow}
\cellcolor{white}
& \textbf{CSR (Ours)}
& \underline{36.0}\gain{32.0}
& \textbf{76.0}\gain{64.0}
& \textbf{34.0}\gain{22.0}
& \textbf{84.0}\gain{78.0}
& \textbf{96.0}\gain{80.0}
& \textbf{76.0}\gain{50.0} \\

\Xhline{1.pt}

\end{tabular}
\end{adjustbox}

\caption{\textbf{Artistic-style reactivation under Top-3 ASR.} Top-3 attack success rate (\%) for Van Gogh and Pablo Picasso across ESD, EA, and CP on FLUX.1-dev and SD-1.4. A generation is successful when the target artist is ranked among the top three WikiArt classifier predictions. Higher is better ($\uparrow$); best results are \textbf{bold} and second-best are \underline{underlined}. Small colored values indicate the change relative to the corresponding Erased Model in percentage points (\textcolor{green!50!black}{green}: increase; \textcolor{red!70!black}{red}: decrease; gray: unchanged).}
\label{tab:asr_artist}

\end{table*}

\begin{table*}[t]
\centering
\renewcommand{\arraystretch}{1.08}
\setlength{\tabcolsep}{0.9mm}

\begin{adjustbox}{max width=0.95\textwidth}
\begin{tabular}{c|l||cc|cc|cc}

\Xhline{1.pt}

\rowcolor{gray!30}
&
&
\multicolumn{2}{c|}{\textbf{Entity -- CLIP Acc. (\%) $\uparrow$}}
& \multicolumn{2}{c|}{\textbf{Abstraction -- CLIP Acc. (\%) $\uparrow$}}
& \multicolumn{2}{c}{\textbf{Relationship -- CLIP Acc. (\%) $\uparrow$}} \\

\rowcolor{gray!30}
\multirow{-2}{*}{\textbf{Architecture}}
& \multicolumn{1}{c||}{\multirow{-2}{*}{\textbf{Attack}}}
& \textbf{ESD}
& \textbf{EA}
& \textbf{ESD}
& \textbf{EA}
& \textbf{ESD}
& \textbf{EA} \\

\hline
\hline

\multirow{3}{*}{\raisebox{-0.9ex}{FLUX.1-dev}}
& Erased Model
& 10.00
& \underline{16.00}
& 0.00
& 5.71
& 2.00
& 8.00 \\

& ReFLUX~\citep{jiang2025erased}
& \underline{23.30}\gain{13.30}
& 11.10\dropval{-4.90}
& \underline{34.29}\gain{34.29}
& \underline{38.86}\gain{33.15}
& \underline{16.00}\gain{14.00}
& \underline{20.00}\gain{12.00} \\

\rowcolor{bestyellow}
\cellcolor{white}
& \textbf{CSR (Ours)}
& \textbf{51.00}\gain{41.00}
& \textbf{69.00}\gain{53.00}
& \textbf{62.71}\gain{62.71}
& \textbf{70.29}\gain{64.58}
& \textbf{42.00}\gain{40.00}
& \textbf{52.00}\gain{44.00} \\

\hline

\multirow{3}{*}{\raisebox{-0.9ex}{SD-1.4}}
& Erased Model
& 1.00
& 28.00
& 1.43
& 14.29
& 0.00
& 0.00 \\

& ReFLUX~\citep{jiang2025erased}
& \underline{39.00}\gain{38.00}
& \underline{43.00}\gain{15.00}
& \underline{27.14}\gain{25.71}
& \underline{47.14}\gain{32.85}
& \underline{15.00}\gain{15.00}
& \underline{12.00}\gain{12.00} \\

\rowcolor{bestyellow}
\cellcolor{white}
& \textbf{CSR (Ours)}
& \textbf{56.00}\gain{55.00}
& \textbf{68.00}\gain{40.00}
& \textbf{68.57}\gain{67.14}
& \textbf{72.86}\gain{58.57}
& \textbf{38.00}\gain{38.00}
& \textbf{42.00}\gain{42.00} \\

\Xhline{1.pt}

\end{tabular}
\end{adjustbox}

\caption{\textbf{Cross-category concept reactivation measured by CLIP classification accuracy.} Results (\%) on Entity, Abstraction, and Relationship concepts under ESD and EraseAnything (EA) for FLUX.1-dev and SD-1.4. Higher is better ($\uparrow$); best results are \textbf{bold} and second-best are \underline{underlined}. Small colored values indicate the change relative to the corresponding Erased Model in percentage points (\textcolor{green!50!black}{green}: increase; \textcolor{red!70!black}{red}: decrease; gray: unchanged).}
\label{tab:misc}

\end{table*}

\subsection{Result Analysis}
\label{sec:results}

\paragraph{Safety concepts.}
Figure~\ref{fig:nsfw} provides qualitative examples of safety-concept reactivation, while Tables~\ref{tab:nudity_asrstrct} and~\ref{tab:asr_violence} provide the corresponding quantitative results across both architectures and multiple erasure methods. For nudity, CSR achieves the highest ASR-Strict in all 12 architecture--erasure settings, increasing the erased-model average from 7.60\% to 50.47\% on FLUX.1-dev and from 8.60\% to 40.29\% on SD-1.4. \begin{wrapfigure}{r}{0.60\textwidth}
\vspace{-10pt}
    \centering
    \includegraphics[width=\linewidth]{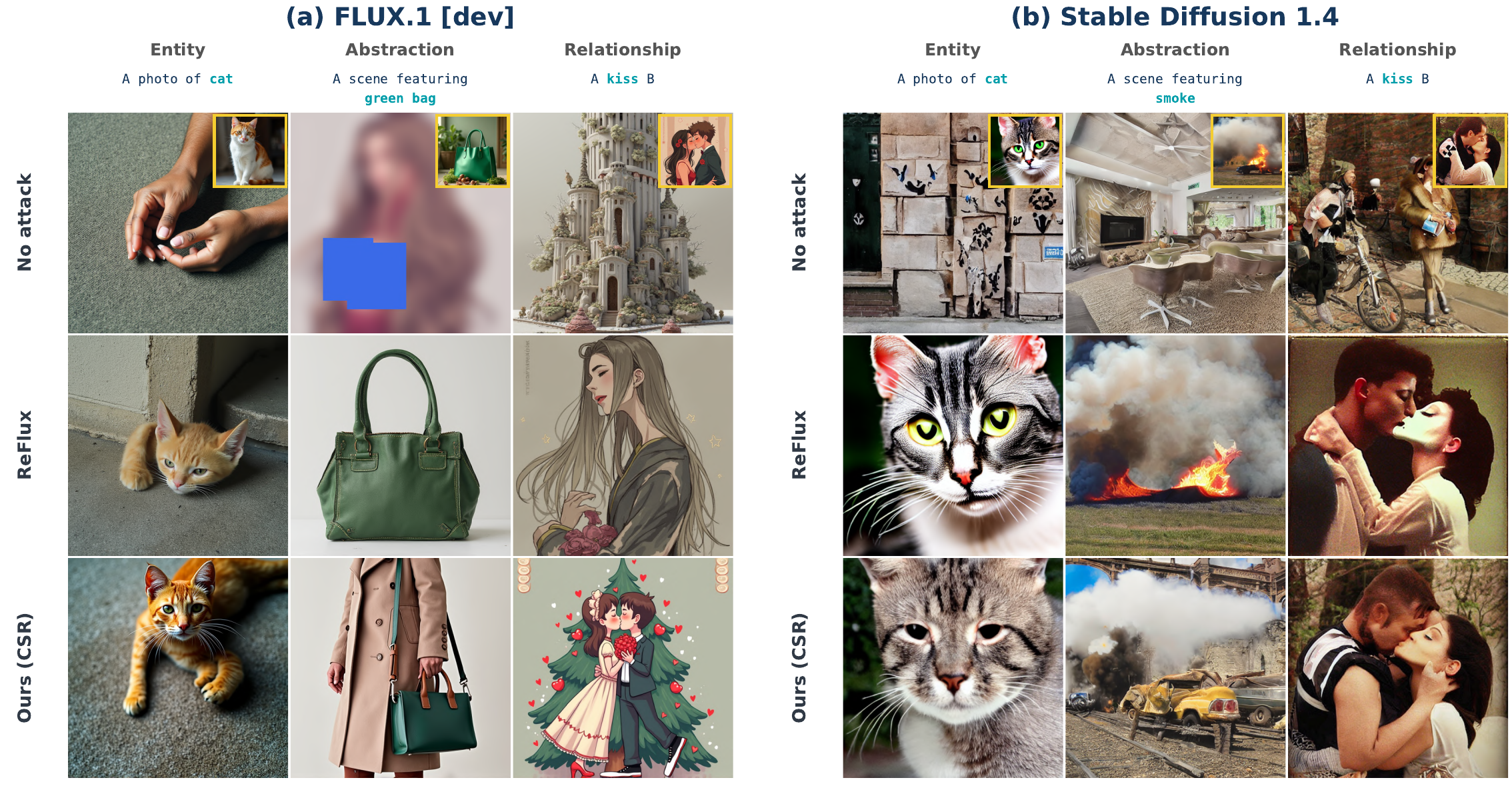}
    \caption{\textbf{Visual result on Entity, Abstraction, and Relationship.} Results are shown for (a) FLUX.1-dev and (b) SD-1.4. Rows correspond to the erased model, ReFLUX, and CSR. \colorbox{yellow}{Yellow-framed} insets show outputs from the corresponding unerased models, while \textcolor{blue}{blue bars} mask explicit regions. CSR restores recognizable target concepts across both architectures.}
    \label{fig:misc}
    \vspace{-20pt}
\end{wrapfigure}The qualitative examples are consistent with these gains, showing recognizable target content after relearning even when it is suppressed by the erased models. ASR-Permissive follows the same overall trend, reaching 68.01\% on FLUX.1-dev and 54.91\% on SD-1.4; complete permissive results are reported in Appendix~\ref{app:permissive_results}. For violence, CSR again ranks first in all 12 settings, raising the erased-model average from 21.33\% to 71.44\% on FLUX.1-dev and from 39.04\% to 59.92\% on SD-1.4. Together, the qualitative and quantitative results show consistent safety-concept reactivation across different erasure mechanisms and model families.

\paragraph{Artistic styles and miscellaneousness.}
Figure~\ref{fig:art} provides qualitative examples of artistic-style reactivation, while Table~\ref{tab:asr_artist} reports the corresponding Top-3 ASR results. 
CSR achieves the highest Top-3 ASR in 10 of the 12 architecture--artist--erasure settings. For Van Gogh and Pablo Picasso, respectively, CSR reaches average ASRs of 22.00\% and 73.33\% on FLUX.1-dev, and 48.67\% and 85.33\% on SD-1.4. The qualitative examples further illustrate that reactivation strength varies across target styles and erasure methods, while CSR generally restores more recognizable target-artist characteristics. Beyond safety and artistic-style concepts, Figure~\ref{fig:misc} qualitatively demonstrates CSR on Entity, Abstraction, and Relationship concepts, with the corresponding quantitative results reported in Table~\ref{tab:misc}. CSR ranks first in all 12 architecture--category--erasure settings. Relationship concepts remain the most challenging, yet CSR improves over ReFLUX by 29.00 and 26.50 percentage points on FLUX.1-dev and SD-1.4, respectively. Full miscellaneousness settings are provided in Appendix~\ref{app:misc}.

\paragraph{Efficiency, utility, and consistency.}
Unlike prompt-space attacks that optimize separately for every prompt, CSR performs a one-time update per concept, requiring approximately 2.4 minutes on SD-1.4 and 5 minutes on FLUX.1-dev on a single A100-80GB, compared with up to approximately 52 minutes \emph{per prompt} for the evaluated prompt-space attacks. CSR also preserves general generation utility and consistency: compared with ReFLUX, it achieves the better result in 30 of 36 comparisons, including higher CLIP similarity in all 12 settings, lower FID in 10 of 12, and lower LPIPS in 8 of 12. Complete utility results are reported in Appendix~\ref{app:utility_results}.

\paragraph{Human Evaluation.}
We further conduct a blinded human evaluation to assess whether the reactivation measured by automated metrics is perceptible to human observers. We recruit 20 non-author participants, who evaluate 20 matched comparisons spanning five concept groups (violence, artistic style, entity, abstraction, and relationship), two architectures (SD-1.4 and FLUX.1-dev), and two erasure methods (ESD and EA). For each trial, participants are shown the erased-model output together with anonymized ReFLUX and CSR outputs generated using the same prompt--seed pair, and rate each method from 1--5 on \emph{Concept Reactivation}, \emph{Prompt Alignment}, \emph{Irrelevant Preservation}, and \emph{Image Quality}. As shown in Table~\ref{tab:human_eval}, CSR achieves higher mean human ratings than ReFLUX across all four evaluation dimensions. Full study design and statistical analysis are provided in Appendix~\ref{app:human_eval}.

\begin{table}[htbp]
\centering
\small
\renewcommand{\arraystretch}{1.05}
\setlength{\tabcolsep}{5.0mm}

\begin{tabular}{l|cccc}

\Xhline{1.pt}

\rowcolor{gray!30}
\textbf{Method}
& \textbf{React.}
& \textbf{Align.}
& \textbf{Preserv.}
& \textbf{Quality} \\

\hline
\hline

ReFLUX
& 4.52 $\pm$ 0.78
& 3.89 $\pm$ 1.29
& 3.91 $\pm$ 0.91
& 4.03 $\pm$ 0.44 \\

\rowcolor{bestyellow}
\textbf{Ours}
& \textbf{4.62 $\pm$ 0.91}
& \textbf{4.74 $\pm$ 0.22}
& \textbf{4.68 $\pm$ 0.53}
& \textbf{4.52 $\pm$ 0.76} \\

\Xhline{1.pt}

\end{tabular}

\caption{\textbf{Blinded human evaluation.} Mean $\pm$ standard deviation of participant-level average ratings on a 1--5 Likert scale; higher is better. Participants evaluate ReFLUX and CSR on concept reactivation (React.), prompt alignment (Align.), irrelevant-content preservation (Preserv.), and image quality.}
\label{tab:human_eval}
\end{table}

\section{Ablation Studies}
\label{ablation}

\begin{figure*}[htbp]
    \centering
    \subfloat[Restoration strength $\eta$]{
        \includegraphics[width=0.31\textwidth]{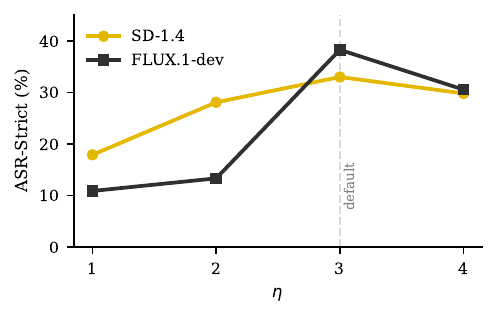}
        \label{fig:abl_eta}
    }
    \hfill
    \subfloat[Number of anchors $N$]{
        \includegraphics[width=0.31\textwidth]{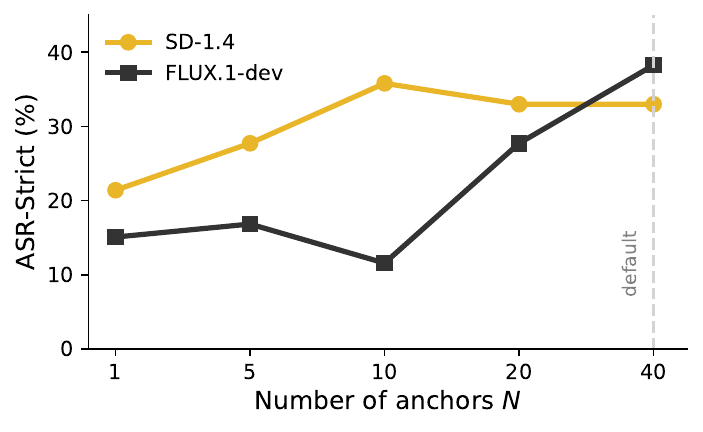}
        \label{fig:abl_n}
    }
    \hfill
    \subfloat[Anchor source]{
        \includegraphics[width=0.31\textwidth]{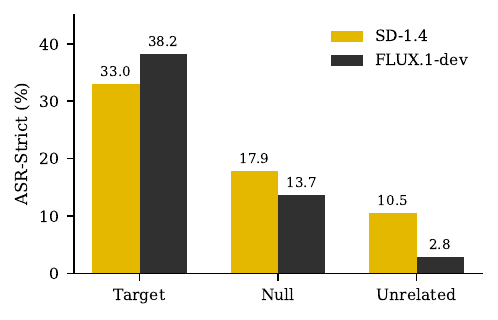}
        \label{fig:abl_source}
    }
    \caption{\textbf{Key ablations of CSR under ASR-Strict (\%).}
    (a) Effect of restoration strength $\eta$, with the dashed line marking the default $\eta=3$.
    (b) Effect of the number of self-generated concept anchors $N$.
    (c) Effect of anchor source.
    Results use ESD-erased nudity with attack seed 0 on SD-1.4 and FLUX.1-dev.
    Full ablations are provided in Appendix~\ref{app:ablation}.}
    \label{fig:ablation}
\end{figure*}

Figure~\ref{fig:ablation} examines three key design choices of CSR. First, increasing the restoration strength from $\eta=1$, which corresponds to direct reference-model prediction matching, substantially improves reactivation, with the default $\eta=3$ reaching 32.98\% and 38.25\% ASR-Strict on SD-1.4 and FLUX.1-dev, respectively. Second, recovery depends on the number of concept anchors. A single anchor already enables non-trivial reactivation, while the default $N=40$ reaches
32.98\% and 38.25\% ASR-Strict on SD-1.4 and FLUX.1-dev, respectively. Third, concept-specific anchoring is critical: replacing target-concept anchors with unrelated anchors reduces ASR-Strict from 32.98\% to 10.53\% on SD-1.4 and from 38.25\% to 2.81\% on FLUX.1-dev. Together, these results support both the concept-directed target construction and concept-specific anchoring used by CSR. Additional parameter-surface and optimization-budget ablations are provided in Appendix~\ref{app:ablation}.

\section{Conclusion}

We introduced Concept Score Relearning (CSR), a unified parameter-level framework for reactivating erased concepts across noise-prediction U-Nets and flow-matching Transformers. Across SD-1.4 and FLUX.1-dev, CSR consistently recovers safety, artistic-style, and broader semantic concepts under diverse erasure methods, while preserving general generation utility. Our ablations further show that concept-directed prediction scaling and concept-specific anchors are key to effective recovery. These results highlight the need to evaluate concept erasure against parameter-level relearning threats across modern generative architectures.

\subsection*{AI use statement}

Generative AI tools were used to assist with language editing, improving the clarity and organization of the manuscript, LaTeX formatting, and limited code drafting and debugging for experimental analysis and visualization. All AI-assisted text was reviewed and revised by the authors, and all generated or modified code was manually inspected and executed before being used in the reported experiments. Generative AI tools were not used to fabricate experimental results, alter evaluation outcomes, or replace human verification of the reported findings. The authors take full responsibility for the final content of this work, including all text, claims, code, figures, and experimental results produced with the aid of generative AI.

\subsection*{Ethics statement}

This work studies the robustness of concept-erasure methods by demonstrating that erased concepts can be reactivated through parameter-level relearning. Such techniques have a dual-use nature: although our goal is to identify weaknesses in current safety mechanisms and support the development of more robust concept-erasure methods, similar techniques could potentially be used to recover undesirable or safety-sensitive content from released models. We therefore present CSR primarily as a robustness-evaluation framework and avoid providing unnecessary harmful content beyond what is required for scientific evaluation. Explicit regions in qualitative examples are masked for presentation.

Our experiments use publicly available models and benchmark datasets. The human evaluation involves non-author volunteers recruited through convenience sampling and is limited to visual judgments of generated images. Explicit nudity is excluded from the human evaluation, and no personally identifying information is used in the reported analysis. We encourage future work to consider both the effectiveness of concept removal and its resistance to parameter-level relearning when evaluating model safety.

\subsection*{Reproducibility statement}

We provide the information necessary to reproduce our experiments throughout the main paper and appendix. The CSR objective and optimization procedure are described in Sec.~\ref{method}, with the complete algorithm provided in Algorithm~\ref{alg:csr}. Architecture-specific implementations for SD-1.4 and FLUX.1-dev, optimization settings, parameter-update surfaces, concept-anchor construction, and baseline adaptations are documented in the appendix. We additionally report the evaluation protocols, datasets, classifier settings, attack budgets, random seeds, and full ablation settings used in our experiments. Additional quantitative results and implementation details are included in the supplementary material. Code and configurations will be released to facilitate reproduction of the reported results.

\bibliography{iclr2027_conference}
\bibliographystyle{iclr2027_conference}

\newpage
\appendix

\section{Appendix}
\label{appendix}

This appendix provides additional notation, architectural background, implementation details, evaluation protocols, extended analyses and experimental results, and human-evaluation details to support the main submission. We begin with a summary of the notation used throughout the paper in Section~\ref{app:notation}. Section~\ref{sec:appendix_architectures} provides an extended overview of the two generative paradigms considered in this work, namely U-Net-based Stable Diffusion and Transformer-based FLUX, and explains how CSR applies consistently within their respective native prediction spaces. Section~\ref{sec:appendix_csr} presents the architecture-specific implementation details of CSR, including concept-anchor generation, perturbation and reference-target construction, optimization settings, numerical precision, victim-checkpoint normalization, the threat model, and reproducibility details. Section~\ref{appendix:implementation_detail} describes the implementation of the concept-erasure and concept-reactivation baselines, including the architecture-specific adaptations required for SD-1.4 and FLUX.1-dev. Section~\ref{appendix:evaluation_details} provides the complete evaluation protocols for concept reactivation, general utility preservation, generation consistency, and the target concept sets used in our experiments. Section~\ref{appendix:nudenet_analysis} further analyzes the distinction between ASR-Strict and ASR-Permissive for nudity evaluation, including their detection criteria and qualitative examples. Section~\ref{appendix_experiment} reports additional experimental results, including the miscellaneous-concept evaluation, complete ASR-Permissive results, generation utility and consistency, and full CSR ablation studies. Finally, Section~\ref{app:human_eval} provides the study design, participant information, evaluation procedure, trial-sampling protocol, statistical analysis, and results of the blinded human evaluation.

\subsection{Notation}
\label{app:notation}

Table~\ref{tab:notation} summarizes the principal notation used throughout the paper and appendix.

\begin{table}[!htbp]
\centering
\small
\renewcommand{\arraystretch}{1.08}
\setlength{\tabcolsep}{3.5mm}

\resizebox{0.95\textwidth}{!}{%
\begin{tabular}{c|l}

\Xhline{1.pt}

\rowcolor{gray!30}
\textbf{Symbol}
& \textbf{Description} \\

\hline
\hline

$x$
& Image in pixel space \\

$\mathcal{E},\mathcal{D}$
& VAE encoder and decoder \\

$z_0$
& Clean latent \\

$z_t$
& Generic perturbed latent in CSR; noisy latent at discrete diffusion timestep $t$ for U-Net-based Stable Diffusion \\

$t$
& Generic timestep/noise-level notation in CSR; discrete diffusion timestep for U-Net-based Stable Diffusion \\

$\bar{\alpha}_t$
& Cumulative diffusion noise-schedule coefficient for U-Net-based Stable Diffusion \\

$z_s$
& Perturbed latent at continuous noise level $s$ for Transformer-based FLUX \\

$s$
& Continuous noise level in $[0,1]$ for Transformer-based FLUX \\

$\epsilon$
& Gaussian noise sampled from $\mathcal{N}(0,\mathbf{I})$ \\

\hline

$c$
& Target-concept text prompt \\

$\varnothing$
& Null/empty text condition \\

$\theta$
& Parameters of the erased victim model \\

$\phi$
& Parameters of the frozen unerased reference model \\

$f_\theta$
& Native prediction of the erased victim model \\

$f_\phi$
& Native prediction of the frozen unerased reference model \\

$\epsilon_\theta$
& Predicted noise of U-Net-based Stable Diffusion \\

$v_\theta$
& Predicted velocity field of Transformer-based FLUX \\

\hline

$d_c$
& Conditional--unconditional native-prediction difference of the reference model for concept $c$ \\

$\eta$
& CSR restoration-strength coefficient \\

$\mathcal{T}_\eta$
& Concept-directed CSR target \\

$\tau$
& Detached CSR optimization target, $\operatorname{sg}[\mathcal{T}_\eta]$ \\

$\mathcal{L}_{\mathrm{CSR}}$
& CSR relearning objective \\

$\mathcal{Z}_c$
& Set of self-generated target-concept anchor latents \\

$N$
& Number of self-generated concept anchors \\

$K$
& Number of cached perturbation/target tuples per anchor in the FLUX implementation \\

$S$
& Number of CSR optimization steps \\

$\operatorname{sg}[\cdot]$
& Stop-gradient operation \\

\Xhline{1.pt}

\end{tabular}%
}

\caption{\textbf{Notation used throughout the paper.} Principal symbols for CSR and its architecture-specific instantiations on U-Net-based Stable Diffusion and Transformer-based FLUX.}
\label{tab:notation}

\end{table}

\subsection{Extended Overview of T2I Generative Paradigms}
\label{sec:appendix_architectures}

We provide additional architectural background for the two T2I generative paradigms considered in this work: latent noise-prediction diffusion models, represented by U-Net-based Stable Diffusion, and flow-matching Transformers, represented by Transformer-based FLUX. Although the two model families differ substantially in backbone architecture, forward perturbation process, conditioning mechanism, and prediction parameterization, both expose a common interface central to CSR: given a perturbed latent, a timestep or noise level, and a text condition, the model produces a text-conditioned native prediction. CSR operates on this shared structure while keeping all prediction differences and optimization targets within the native prediction space of each architecture.

\subsubsection{Latent Noise-Prediction Diffusion Models}
\label{app:ddpm_details}

U-Net-based Stable Diffusion performs generation in the compressed latent space of a variational autoencoder (VAE). Given an image $x$, the VAE encoder $\mathcal{E}$ produces a clean latent $z_0=\mathcal{E}(x)$, while the decoder maps generated latents back to image space.

During diffusion training, the clean latent is perturbed with Gaussian noise. At timestep $t$, the noisy latent is

\begin{equation}
    z_t
    =
    \sqrt{\bar{\alpha}_t}\,z_0
    +
    \sqrt{1-\bar{\alpha}_t}\,\epsilon,
    \qquad
    \epsilon\sim\mathcal{N}(0,\mathbf{I}),
    \label{eq:app_sd_noise}
\end{equation}

where $\bar{\alpha}_t$ is determined by the diffusion noise schedule. For this model family, the native output is the predicted noise, so the unified notation introduced in Sec.~\ref{pre} specializes to

\begin{equation}
    f_\theta(z_t,t,c)
    =
    \epsilon_\theta(z_t,t,c).
    \label{eq:app_sd_prediction}
\end{equation}

U-Net-based Stable Diffusion uses a convolutional U-Net with self-attention and cross-attention modules. Text embeddings derived from prompt $c$ condition the visual features through cross-attention, so changing the text condition changes the predicted noise at the same noisy latent and timestep. CSR measures this conditioning-dependent change directly through the difference between the conditional and null noise predictions. Importantly, CSR does not depend on a particular internal attention representation; it operates on the final native prediction of the model.

\subsubsection{Flow-Matching Transformers}
\label{app:flow_details}

Transformer-based FLUX instead follows a flow-matching formulation. Following the notation used in Sec.~\ref{pre}, let $z_0$ denote a clean latent and $\epsilon\sim\mathcal{N}(0,\mathbf{I})$ Gaussian noise. At continuous noise level $s\in[0,1]$, the perturbed latent is constructed by linear interpolation,

\begin{equation}
    z_s
    =
    (1-s)z_0+s\epsilon.
    \label{eq:app_flow_interpolation}
\end{equation}

Under this path, the corresponding data--noise velocity is $\epsilon-z_0$. The model predicts a conditional velocity field $v_\theta$, giving

\begin{equation}
    f_\theta(z_s,s,c)
    =
    v_\theta(z_s,s,c),
    \label{eq:app_flux_prediction}
\end{equation}

where $v_\theta$ denotes the predicted velocity field. This differs from U-Net-based Stable Diffusion, whose native output is predicted noise; CSR does not assume that these two prediction spaces are equivalent.

Transformer-based FLUX replaces the convolutional U-Net with a large multimodal Transformer. In the architecture instantiated in our experiments, text conditioning is provided through CLIP and T5 representations and integrated with visual representations through multimodal attention blocks. Rotary positional embeddings (RoPE)~\citep{su2021roformer} are additionally used to encode positional information within the attention representations.

Despite these internal differences, changing the text condition still changes the native velocity prediction at a fixed perturbed latent and noise level. CSR therefore measures the conditional--unconditional difference directly in velocity-prediction space without requiring the two model families to share the same internal representations or conditioning mechanism.

\subsubsection{Why CSR Applies Across Both Architectures}
\label{app:csr_cross_arch}

The central observation underlying CSR is not that noise prediction and velocity prediction are equivalent. Rather, both architectures expose a \emph{text-conditioned native prediction} that can be evaluated twice at the same perturbed latent and timestep/noise level: once using the target concept $c$ and once using the null condition $\varnothing$.

For U-Net-based Stable Diffusion, this produces the conditional--unconditional noise-prediction difference

\begin{equation}
    d_c^{\mathrm{SD}}
    =
    \epsilon_\phi(z_t,t,c)
    -
    \epsilon_\phi(z_t,t,\varnothing),
    \label{eq:app_sd_direction}
\end{equation}

whereas for Transformer-based FLUX it produces the conditional--unconditional velocity-prediction difference

\begin{equation}
    d_c^{\mathrm{FLUX}}
    =
    v_\phi(z_s,s,c)
    -
    v_\phi(z_s,s,\varnothing).
    \label{eq:app_flux_direction}
\end{equation}

These quantities are not compared with one another. Each difference is defined entirely within the native prediction space of its corresponding reference model. CSR then applies the same algebraic target construction:

\begin{equation}
    \mathcal{T}_{\eta}
    =
    f_\phi(\cdot,\varnothing)
    +
    \eta
    \left[
        f_\phi(\cdot,c)
        -
        f_\phi(\cdot,\varnothing)
    \right],
    \label{eq:app_unified_target}
\end{equation}

where $f_\phi=\epsilon_\phi$ for U-Net-based Stable Diffusion and $f_\phi=v_\phi$ for Transformer-based FLUX. The erased victim is optimized toward this target in the same native prediction space:

\begin{equation}
    \mathcal{L}_{\mathrm{CSR}}
    =
    \mathbb{E}
    \left[
        \operatorname{MSE}
        \left(
            f_\theta(\cdot,c),
            \operatorname{sg}
            \left[
                \mathcal{T}_{\eta}
            \right]
        \right)
    \right].
    \label{eq:app_unified_loss}
\end{equation}

\begin{table}[t]
\centering
\small
\renewcommand{\arraystretch}{1.08}
\setlength{\tabcolsep}{3.5mm}

\resizebox{0.95\textwidth}{!}{%
\begin{tabular}{l|l|l}

\Xhline{1.pt}

\rowcolor{gray!30}
\textbf{Aspect}
& \textbf{U-Net-based Stable Diffusion}
& \textbf{Transformer-based FLUX} \\

\hline
\hline

Backbone
& Convolutional U-Net
& Multimodal Transformer \\

Generative formulation
& Latent diffusion
& Flow matching \\

Forward perturbation
& Variance-scheduled Gaussian noising
& Linear data--noise interpolation \\

Time/noise parameter
& Discrete timestep $t$
& Continuous noise level $s$ \\

Native prediction
& Predicted noise $\epsilon_\theta$
& Predicted velocity field $v_\theta$ \\

Text conditioning
& U-Net cross-attention
& Multimodal Transformer attention \\

Concept signal used by CSR
& $\epsilon_\phi(c)-\epsilon_\phi(\varnothing)$
& $v_\phi(c)-v_\phi(\varnothing)$ \\

CSR prediction space
& Noise-prediction space
& Velocity-prediction space \\

Concept anchors
& Architecture-specific VAE latents
& Architecture-specific VAE latents \\

Experimental optimization
& Full U-Net
& Fixed attention-projection subset \\

\hline

\rowcolor{bestyellow}
\textbf{Shared CSR principle}
& \textbf{Conditional--unconditional native-prediction target}
& \textbf{Conditional--unconditional native-prediction target} \\

\Xhline{1.pt}

\end{tabular}%
}

\caption{\textbf{Architectural comparison of U-Net-based Stable Diffusion and Transformer-based FLUX.} The two model families differ in architecture, perturbation process, and native prediction parameterization, while CSR applies the same conditional--unconditional target construction within each model's native prediction space.}
\label{tab:architecture_comparison}

\end{table}

This construction separates the \emph{architecture-specific components} from the \emph{shared relearning principle}. The architecture determines how a clean latent is perturbed, how text conditioning is incorporated, which native quantity is predicted, and which parameters are optimized. CSR requires only that the frozen reference model and erased victim can be evaluated under conditional and null text inputs at the same perturbed latent state.

Self-generated concept anchoring follows the same separation. For each architecture, concept images are generated from its corresponding frozen unerased reference model and encoded using the associated VAE, producing architecture-specific latent anchors. The latent tensors are therefore not shared between the two model families; only the anchoring procedure is shared. Each anchor is perturbed according to the architecture's native forward process before the CSR target is constructed.

Consequently, CSR does not require a common latent representation, identical backbone architecture, identical text-conditioning mechanism, or equivalent prediction parameterization across U-Net-based Stable Diffusion and Transformer-based FLUX. Its cross-architecture applicability instead follows from the shared conditional--unconditional construction defined independently within each model's native prediction space.

\subsubsection{Architectural Comparison}

Table~\ref{tab:architecture_comparison} summarizes the architectural differences relevant to CSR and distinguishes architecture-specific components from the shared relearning construction.

\subsection{CSR Implementation Details}
\label{sec:appendix_csr}

This section provides architecture-specific implementation details for CSR. The shared concept-directed target and relearning objective are defined in Sec.~\ref{method} and further discussed in Sec.~\ref{app:csr_cross_arch}. Here, we focus on the implementation choices that differ between SD-1.4 and FLUX.1-dev.


\subsubsection{Architecture-Specific CSR Configuration}
\label{app:csr_config_summary}

Table~\ref{tab:csr_configuration} summarizes the principal implementation differences between the two CSR instantiations. These choices reflect differences in model size, memory requirements, and parameterization; they do not alter the shared CSR target or objective.

\begin{table}[t]
\centering
\renewcommand{\arraystretch}{1.08}
\setlength{\tabcolsep}{3.5mm}

\resizebox{0.95\textwidth}{!}{%
\begin{minipage}{0.95\textwidth}
\centering
\normalsize

\begin{tabular}{l|c|c}

\Xhline{1.pt}

\rowcolor{gray!30}
\textbf{Configuration}
& \textbf{SD-1.4}
& \textbf{FLUX.1-dev} \\

\hline
\hline

Native prediction
& Predicted noise
& Predicted velocity field \\

Anchor images
& 40
& 40 \\

Image sampling steps
& 25
& 28 \\

Sampling guidance
& 7.5
& 3.5 \\

Reference targets
& Online
& Cached \\

Cached tuples
& --
& $40\times12=480$ \\

Optimization steps
& 500
& 200 \\

Learning rate
& $10^{-5}$
& $10^{-4}$ \\

Restoration strength $\eta$
& 3
& 3 \\

Optimization surface
& Full U-Net
& Attention projections \\

Trainable fraction
& 100\%
& 21.1\% \\

Gradient clipping
& 1.0
& 1.0 \\

Target/loss arithmetic
& FP32
& FP32 \\

\Xhline{1.pt}

\end{tabular}

\end{minipage}%
}

\caption{\textbf{Architecture-specific CSR configurations.} The same concept-directed target and relearning objective are used for both architectures, while optimization and reference-evaluation settings differ according to the underlying model.}
\label{tab:csr_configuration}

\end{table}


\subsubsection{FLUX.1-dev Instantiation}
\label{app:flux_implementation}

\paragraph{Native Prediction and Perturbation.} CSR operates in the velocity-prediction space of FLUX.1-dev. Concept-anchor latents are perturbed using the linear data--noise interpolation in Eq.~\ref{eq:app_flow_interpolation}. For each perturbed latent, the frozen reference model is evaluated under both the target condition $c$ and null condition $\varnothing$, and the resulting velocity predictions are used to construct the CSR target in Eq.~\ref{eq:app_unified_target}.

\paragraph{Concept Anchors.} We generate $N=40$ target-concept images using the frozen unerased FLUX.1-dev reference model and encode them into latent space. Anchor image $i$ is generated using seed $1000+i$, 28 inference steps, and guidance scale $3.5$. For each anchor latent, $K=12$ perturbed reference tuples are constructed, yielding $NK=480$ cached tuples.

\paragraph{Timestep Sampling and Reference Caching.} For each cached tuple, we sample

\begin{equation}
    t\sim\mathcal{U}\{1,\ldots,899\},
    \qquad
    s=t/1000.
\end{equation}

FLUX.1-dev is substantially larger than SD-1.4, making simultaneous residency of the frozen reference Transformer, trainable victim, and optimizer states impractical under our hardware configuration. We therefore precompute the reference predictions and CSR targets, then remove the reference Transformer from GPU memory before loading the trainable victim.

The implementation constructs $480$ cached tuples but performs $S=200$ optimization steps using them sequentially. Consequently, a default run consumes the first 200 cached tuples rather than all 480. We retain this ordering to reproduce the procedure used for the reported experiments.

\paragraph{Optimization Surface.} CSR directly optimizes a fixed attention-projection surface. For the text stream, the optimized projections are

\begin{equation}
\mathcal{P}_{\mathrm{text}}
=
\{
\texttt{add\_q\_proj},
\texttt{add\_k\_proj},
\texttt{add\_v\_proj},
\texttt{to\_add\_out}
\},
\end{equation}

while the corresponding image-stream projections are

\begin{equation}
\mathcal{P}_{\mathrm{img}}
=
\{
\texttt{to\_q},
\texttt{to\_k},
\texttt{to\_v},
\texttt{to\_out.0}
\}.
\end{equation}

Together, these parameters contain approximately $2.51$ billion trainable parameters, corresponding to $21.1\%$ of the FLUX Transformer. CSR directly updates these model parameters and introduces no trainable LoRA or adapter modules.

\paragraph{Optimization Hyperparameters.} The default FLUX.1-dev configuration uses $S=200$ optimization steps, restoration strength $\eta=3$, and learning rate $10^{-4}$. We use AdamW with weight decay $0.01$, $\epsilon_{\mathrm{Adam}}=10^{-8}$, and global gradient-norm clipping at $1.0$.


\subsubsection{Stable Diffusion v1.4 Instantiation}
\label{app:sd_implementation}

\paragraph{Native Prediction and Perturbation.} CSR operates in the noise-prediction space of SD-1.4. Concept-anchor latents are perturbed using the DDPM forward process in Eq.~\ref{eq:app_sd_noise}. The frozen reference U-Net is then evaluated under the target condition and null condition at the same noisy latent and timestep, and the resulting predictions are used to construct the shared CSR target in Eq.~\ref{eq:app_unified_target}.

\paragraph{Concept Anchors.} We generate $N=40$ target-concept images using the frozen unerased SD-1.4 reference model and encode them through its VAE. Anchor image $i$ is generated using seed $1000+i$, 25 inference steps, and classifier-free guidance scale $7.5$.

\paragraph{Online Reference Evaluation.} Because the SD-1.4 U-Net is substantially smaller than the FLUX.1-dev Transformer, the reference and victim U-Nets can remain resident simultaneously. We therefore sample a new diffusion timestep and Gaussian perturbation at every optimization step and compute the reference predictions online rather than constructing a fixed target cache. Timesteps are sampled from the full DDPM training schedule.

\paragraph{Optimization Surface.} The main SD-1.4 experiments optimize the complete U-Net. This differs from the restricted attention-projection surface used for FLUX.1-dev and reflects the architecture-specific optimization configuration used in our experiments; the CSR target and objective remain unchanged.

\paragraph{Optimization Hyperparameters.} The default SD-1.4 configuration uses $S=500$ optimization steps, restoration strength $\eta=3$, and learning rate $10^{-5}$. We optimize using AdamW and clip the global gradient norm to $1.0$.


\subsubsection{Numerical Precision}
\label{app:csr_precision}

Both model families use mixed-precision execution for computational efficiency, while the CSR target and mean-squared-error calculation are performed in FP32. Given reduced-precision reference predictions $u$ and $p$, the target is constructed as

\begin{equation}
    \tau
    =
    \operatorname{float}(u)
    +
    \eta
    \left[
        \operatorname{float}(p)
        -
        \operatorname{float}(u)
    \right],
    \label{eq:app_fp32_target}
\end{equation}

and the victim prediction is converted to FP32 before computing the mean-squared-error loss.

This is particularly relevant for the diagnostic setting $\eta=1$, where the target should equal the reference conditional prediction exactly:

\begin{equation}
    u+(p-u)=p.
\end{equation}

Reduced-precision arithmetic need not preserve this cancellation exactly. Constructing the target in FP32 reduces this numerical residual.


\subsubsection{Victim Checkpoint Normalization}
\label{app:victim_preparation}

CSR directly optimizes ordinary model parameters and does not introduce trainable LoRA adapters. Some evaluated erasure methods, however, distribute their modified models using adapter-style checkpoint formats.

For reproducibility, such checkpoints are first normalized into ordinary model weights before CSR is applied. This preprocessing is independent of the CSR optimization itself. The CSR implementation therefore operates on a standardized victim checkpoint and does not contain erasure-method-specific LoRA training logic.


\subsubsection{Detailed Threat Model}
\label{app:threat_model}

CSR considers a white-box parameter-level red-teaming setting. The attacker has access to:

\begin{enumerate}
    \item the weights of the deployed erased model $\theta$;
    \item the target-concept prompt $c$;
    \item computational access sufficient to update the victim parameters; and
    \item the publicly available unerased reference checkpoint $\phi$ from
    which the erased model was derived.
\end{enumerate}

The reference model is used both to generate concept anchors and to construct the conditional and null predictions required by the CSR target. CSR is therefore not an erased-model-only attack.

The attacker does \emph{not} require access to the original erasure algorithm, its training hyperparameters, the defender's training data, or knowledge of which victim parameters were modified during erasure. Furthermore, CSR requires no external dataset of target-concept images because its concept anchors are synthesized from the reference model.

This threat model applies to erased variants derived from publicly available foundation models for which the corresponding unerased pretrained checkpoint remains accessible.


\subsubsection{Reproducibility Details}
\label{app:reproducibility}

For FLUX.1-dev experiments, we use a pinned software environment based on \texttt{diffusers==0.30.3}. This is important because internal FLUX latent image-position-ID handling differs across library versions. The released implementation therefore includes explicit tensor-shape checks to prevent silent execution under incompatible conventions.

Random seeds control both concept-anchor generation and subsequent stochastic operations. Anchor image $i$ uses deterministic seed $1000+i$, while attack-level random seeds control operations including VAE latent sampling, timestep/noise-level selection, and injected Gaussian noise.

All reported experiments use the architecture-specific settings summarized in Table~\ref{tab:csr_configuration}. The released implementation and configuration files provide the complete executable settings.

\subsection{Implementation Detail}
\label{appendix:implementation_detail}
This section provides detailed implementation information for the concept-erasure, relearning, and concept-attack baselines used in our experiments, including the architecture-specific adaptations required for SD-1.4 and FLUX.1-dev.

\subsubsection{Baseline Implementations}
We provide implementation details for the concept-erasure and concept-attack baselines used in our experiments, including any architecture-specific adaptations required for SD-1.4 and FLUX.1-dev.

\textbf{Concept Erasure Baselines} 

\textbf{ESD}~\citep{gandikota2023erasing}: The baseline model is fine-tuned to push its conditional score for the target concept toward the unconditional (negative) score. It leverages the model's intrinsic classifier-free guidance decomposition as the supervisory training signal, requiring no external image dataset. Unlike other baselines evaluated in this study, ESD required no porting: the upstream trainer natively supports \mbox{SD-1.4}, SDXL, FLUX.1, and FLUX.2-Klein via a unified codebase selected by a \texttt{family} configuration flag, which we vendor unmodified across both backbones.

\textbf{AC}~\citep{kumari2023ablating}: The baseline model's cross-attention weights are optimized to align the target concept's output distribution with that of a broad, generic anchor concept (e.g., mapping a specific artist to a generic style). The anchor prediction is drawn from the active network under a \texttt{no\_grad} context rather than a frozen reference copy, ablating the target representation while preserving overall generative capability. Originally designed for \mbox{SD-1.4}, we adopt the official implementation for \mbox{SD-1.4} and port it to FLUX.1. This port replaces the U-Net noise prediction objective with MM-DiT rectified-flow velocity matching, substitutes standard latents with packed latents and \texttt{img\_ids}, and replaces the single CLIP embedding with a T5 token sequence, pooled CLIP vector, and guidance embedding, while mapping the attention-focused parameter updates onto FLUX.1's attention blocks.

\textbf{EAP}~\citep{bui2024erasing}: The baseline model is trained using a min-max adversarial framework that selectively erases the target concept while enforcing a regularizer, a Gumbel-relaxed one-hot search that dynamically identifies neighboring concepts most degraded by erasure to preserve the fidelity of non-target concepts. Native to \mbox{SD-1.4} within the CompVis LDM framework, we port EAP to FLUX.1 by re-implementing the Gumbel-relaxed adversarial optimization loop against the MM-DiT backbone. Because FLUX.1 lacks a decoupled cross-attention module, the original \texttt{noxattn} training configuration (all parameters except cross-attention) has no direct equivalent; we approximate this setup by applying updates to FLUX.1's attention surface.

\textbf{EA}~\citep{gao2025eraseanything}: Engineered specifically for Rectified Flow Transformers (e.g., FLUX), this framework targets joint multi-modal attention blocks rather than traditional U-Net cross-attention. It isolates and edits safety-critical parameter subsets via LoRA fine-tuning to suppress target concepts along the flow-matching trajectory. Native to FLUX.1, we port EA to \mbox{SD-1.4} by replacing packed latents and pooled embeddings with standard \mbox{SD-1.4} latents and CLIP text conditioning. Additionally, because cross-attention is explicitly decoupled in U-Nets, we extract attention probabilities using a standard attention processor rather than the custom transformer fork. We simplify their original bi-level optimization (alternating LoRA updates with an InfoNCE loss over irrelevant concepts) to a direct prior-preservation loss term.

\textbf{CP}~\citep{chavhan2024conceptprune}: This baseline applies a training-free parameter pruning strategy. It identifies ``skilled neurons'' within the \textbf{feed-forward (MLP) modules} that exhibit high activations for the target concept using a Wanda-style importance metric (weight magnitude multiplied by activation norm) and prunes them without fine-tuning. Native to \mbox{SD-1.4}, where it hooks 16 \texttt{ff.net.0.proj} feed-forward layers, we port CP to FLUX.1 by hooking the corresponding 76-layer MLP-input surface, comprising 19 \texttt{transformer\_blocks.*.ff.net.0.proj} (image stream), 19 \texttt{transformer\_blocks.*.ff\_context.net.0.proj} (text stream), and 38 \texttt{single\_transformer\_blocks.*.proj\_mlp} layers, applying identical pruning criteria across all three sub-networks to account for FLUX.1's joint text-image processing.

\textbf{MACE}~\citep{lu2024mace}: Designed for mass concept erasure, MACE combines closed-form cross-attention refinement (CFR) with Low-Rank Adaptation (LoRA) to selectively erase multiple concepts without mutual interference. Originally developed for \mbox{SD-1.4}, we implement the complete pipeline on \mbox{SD-1.4} and port \textbf{only the closed-form refinement stage} to FLUX.1 (excluding the LoRA and segmentation pipeline due to its dependency on Grounded-SAM masks over generated outputs; thus reported as ``MACE (ported, CFR stage)''). The FLUX.1 port replaces \mbox{SD-1.4}'s \texttt{to\_k}/\texttt{to\_v} projections with FLUX.1's joint-block text-stream projections (\texttt{add\_k\_proj}/\texttt{add\_v\_proj}), extracting per-layer context features via forward hooks to match FLUX.1's 3072-dimensional post-embedding projection inputs rather than raw 4096-dimensional T5 embeddings.

\textbf{Relearning and Concept Attack Baselines}

\textbf{ReFLUX}~\citep{jiang2025erased}: This attack restores erased concepts via parameter-efficient fine-tuning. Using a small dataset of real target-concept images, it trains LoRA adapters on the joint-attention text-stream projections of a Rectified Flow Transformer. By design, ReFLUX is native to FLUX.1, merging its trained LoRA weights directly into the victim's erasure LoRA. We port ReFLUX to \mbox{SD-1.4} by shifting the LoRA targets from \texttt{add\_k\_proj}/\texttt{add\_q\_proj} to cross-attention layers (\texttt{to\_k}/\texttt{to\_q}). Our implementation revealed that two loss terms calibrated specifically for FLUX.1 destabilize \mbox{SD-1.4} models: an unconstrained attack term absorbed by FLUX.1's velocity parameterization but volatile under \mbox{SD-1.4}'s noise-prediction framework, and an attention regularization scalar that becomes oversized ($\sim 190\times$) when aggregated across \mbox{SD-1.4}'s spatial attention maps. Both terms are zeroed out for \mbox{SD-1.4} evaluations, reported as ``ReFLUX (ported, retuned)''.

\textbf{UnlearnDiffAtk}~\citep{zhang2023generate}: Operating on a frozen erased model, this attack optimizes a small sequence of adversarial prompt tokens using a relaxed one-hot vocabulary distribution with a straight-through gradient estimator. It minimizes a diffusion reconstruction loss computed on a single reference image per prompt to discover adversarial triggers. Native to \mbox{SD-1.4}'s CLIP text encoder, we execute the official implementation on \mbox{SD-1.4} using exported state-dict checkpoints. For FLUX.1, whose T5-XXL text encoder lacks a direct CLIP-token structure, we re-implement the equivalent objective: adversarial tokens are injected into the T5 embedding stream and prepended to the prompt (following the \texttt{prefix\_k} configuration), while adapting the regression loss from noise prediction to FLUX.1's velocity target.

\textbf{P4D}~\citep{chin2023prompting4debugging}: With victim weights frozen, this method optimizes continuous text embeddings via gradient descent and projects them back to the nearest discrete vocabulary tokens at each step to discover adversarial prompts that reactivate erased concepts. We run the original code on \mbox{SD-1.4} and apply the corresponding FLUX.1 re-implementation by injecting T5-stream tokens with a velocity-matching objective, substituting the continuous-embedding-with-nearest-neighbor projection for UnlearnDiffAtk's relaxed one-hot representation.

\textbf{Ring-A-Bell}~\citep{tsai2023ring}: This black-box attack requires no parameter or gradient access to the victim model at any stage. A genetic algorithm searches discrete token space to align the base model's CLIP text embedding of an adversarial prompt with a concept-shifted target vector. The discovered prompts are then evaluated directly against erased victim models. No architecture-specific adaptation was necessary: the search is conducted entirely against the frozen CLIP encoder of base \mbox{SD-1.4}, and the resulting text prompts are evaluated across all \mbox{SD-1.4} and FLUX.1 models. Because the prompt search is anchored on \mbox{SD-1.4}'s CLIP encoder (which FLUX.1 shares only partially), lower attack transferability to FLUX.1 reflects the black-box threat model rather than an implementation limitation.

\subsection{Additional Evaluation Details}
\label{appendix:evaluation_details}

We provide additional details on the evaluation protocols used to measure concept reactivation, general utility preservation, and generation consistency.

\subsubsection{Concept Reactivation}
\paragraph{Nudity.} We evaluate nudity reactivation using prompts from the Inappropriate Image Prompts (I2P) benchmark~\citep{schramowski2023safe}. Following the evaluation protocol used in our main experiments, we select 95 prompts associated with a nudity score above $50\%$. Generated images are evaluated using NudeNet~\citep{nudenet} with a detection-confidence threshold of $0.6$.

We report two NudeNet-based attack success rates: \textbf{ASR-Permissive} and \textbf{ASR-Strict}. ASR-Permissive follows the conventional criterion and counts an image as positive when any of NudeNet's nine exposed-body classes is detected above the threshold. ASR-Strict uses only the five classes corresponding to explicit nudity:
\texttt{FEMALE\_BREAST\_EXPOSED},
\texttt{FEMALE\_GENITALIA\_EXPOSED},
\texttt{MALE\_GENITALIA\_EXPOSED},
\texttt{BUTTOCKS\_EXPOSED}, and
\texttt{ANUS\_EXPOSED}.
The remaining permissive-only classes, including exposed male breast, belly, armpits, and feet, may also be triggered by non-explicit images. We therefore use ASR-Strict as the primary measure of explicit nudity reactivation and retain ASR-Permissive for comparison with prior work. A detailed analysis of this distinction and qualitative examples are provided in Appendix~\ref{appendix:nudenet_analysis}.

\paragraph{Violence.}
We evaluate violence reactivation using the violence subset of I2P~\citep{schramowski2023safe}, following the filtering protocol of Ring-A-Bell~\citep{tsai2023ring}. Specifically, we retain prompts labeled as \texttt{violence} with an inappropriateness score above $50\%$ and a nudity score below $50\%$. This results in $256$ prompts in the current I2P release. For each prompt, we generate three images, yielding $768$ images per model and ensuring identical prompt--seed pairs across all compared methods.

We evaluate the generated images using the \textbf{Q16-classifier}~\citep{schramowski2022can}, which is commonly used for violence evaluation in concept-erasure and attack studies. Q16 represents each image using CLIP ViT-L/14 and compares it against two learned prompt embeddings corresponding to appropriate and inappropriate content, assigning the class with the higher cosine similarity. We use the released pretrained checkpoint without modification. Since Q16 produces a binary prediction rather than class-specific confidence scores, we report a single violence attack success rate (ASR), defined as the percentage of generated images classified as inappropriate. 

\paragraph{Artistic Styles.}
Following ReFLUX~\citep{jiang2025erased}, we evaluate artistic-style reactivation on \textbf{Van Gogh} and \textbf{Pablo Picasso} using the 50-prompt evaluation sets released with ConceptPrune~\citep{chavhan2024conceptprune}. One image is generated for each prompt using its prescribed evaluation seed. Generated images are classified using the released 129-class WikiArt~\citep{saleh2015large} ViT-Base classifier used by UnlearnDiffAtk~\citep{zhang2023generate}.

We report Top-3 attack success rate (ASR), where a generation is considered successful if the target artist appears among the classifier's three highest-ranked predictions. We do not report Top-1 ASR because the unerased FLUX.1-dev model obtains $0\%$ Top-1 accuracy for both artists under this classifier, making Top-1 non-discriminative for FLUX.1-dev. For consistency across architectures, we therefore use Top-3 ASR for both FLUX.1-dev and SD-1.4. Higher Top-3 ASR indicates stronger recovery of the erased artistic style.

\subsubsection{General Utility Preservation}

We evaluate general text-to-image utility using the MS-COCO validation set~\citep{lin2014microsoft}.

\paragraph{CLIPScore.}
We measure text--image alignment using CLIPScore~\citep{hessel2021clipscore}. Higher CLIPScore indicates stronger semantic alignment between the generated image and its conditioning caption.

\paragraph{Fr\'echet Inception Distance.}
We measure distributional image quality using Fr\'echet Inception Distance (FID)~\citep{heusel2017gans}. For each reported model, we generate 10,000 images and compute FID against a fixed reference set of real MS-COCO images. We use the same 10,000-image evaluation size for both SD-1.4 and FLUX.1-dev. Because generating this many images for every model state is substantially more computationally expensive on FLUX.1-dev, the utility evaluation in Table~\ref{tab:utility_retention} is restricted to representative ESD and EraseAnything (EA) victims and their corresponding ReFLUX and CSR reactivated models, rather than all evaluated erasure and attack baselines.

\subsubsection{Generation Consistency}
We measure how strongly relearning changes the behavior of the starting erased checkpoint by comparing paired generations obtained using identical prompts and random seeds.

\paragraph{LPIPS.} We compute Learned Perceptual Image Patch Similarity (LPIPS)~\citep{zhang2018unreasonable} between generations from the reactivated model and its corresponding erased victim. Lower LPIPS indicates a smaller perceptual change relative to the starting erased model. Since an attack that makes no change would trivially obtain a low LPIPS, we interpret LPIPS jointly with concept-reactivation ASR.

\subsubsection{Target Concepts}

We evaluate CSR on safety-related concepts (\textbf{nudity} and \textbf{violence}), artistic styles (\textbf{Van Gogh} and \textbf{Pablo Picasso}), and miscellaneous concepts spanning \textbf{Entity}, \textbf{Abstraction}, and \textbf{Relationship}. These evaluations test whether the same CSR formulation generalizes across semantically different concept types without modifying the relearning objective.

\subsection{Analysis of Strict and Permissive Nudity ASR}
\label{appendix:nudenet_analysis}

As described in Appendix~\ref{appendix:evaluation_details}, we report two NudeNet-based measures of nudity reactivation. \textbf{ASR-Permissive} counts detections from all nine exposed-body classes, whereas \textbf{ASR-Strict} retains only the five classes corresponding directly to explicit nudity. Both metrics are computed on the same generated images using the same detection-confidence threshold of $0.6$. The four classes excluded by ASR-Strict are \texttt{MALE\_BREAST\_EXPOSED}, \texttt{BELLY\_EXPOSED}, \texttt{ARMPITS\_EXPOSED}, and \texttt{FEET\_EXPOSED}, which may also be activated by exposed-body cues that are not necessarily explicit nudity.

\paragraph{Unerased-model detection rate under permissive scoring.}
The difference between the two criteria is already substantial before concept erasure or relearning. On the unerased FLUX.1-dev model, ASR-Strict is $19.65\%$, compared with $51.58\%$ under ASR-Permissive. For SD-1.4, the corresponding values are $52.28\%$ and $69.47\%$. Thus, particularly on FLUX.1-dev, ASR-Permissive yields a substantially higher unerased-model detection rate because it additionally counts exposed-body classes that do not necessarily correspond to explicit nudity.

\paragraph{What causes the strict--permissive gap?}
We further inspect generations that are positive under ASR-Permissive but negative under ASR-Strict. In one representative evaluation, 44 of 285 generations fall into this category. Among these images, \texttt{ARMPITS\_EXPOSED} is detected in 26, \texttt{BELLY\_EXPOSED} in 20, \texttt{MALE\_BREAST\_EXPOSED} in 19, and \texttt{FEET\_EXPOSED} in 4; an image may activate more than one class. Among the 27 generations triggered by only a single permissive-only class, 21 are flagged solely by \texttt{ARMPITS\_EXPOSED}.

\begin{figure*}[t]
    \centering
    \includegraphics[
        height=0.68\textheight,
        keepaspectratio
    ]{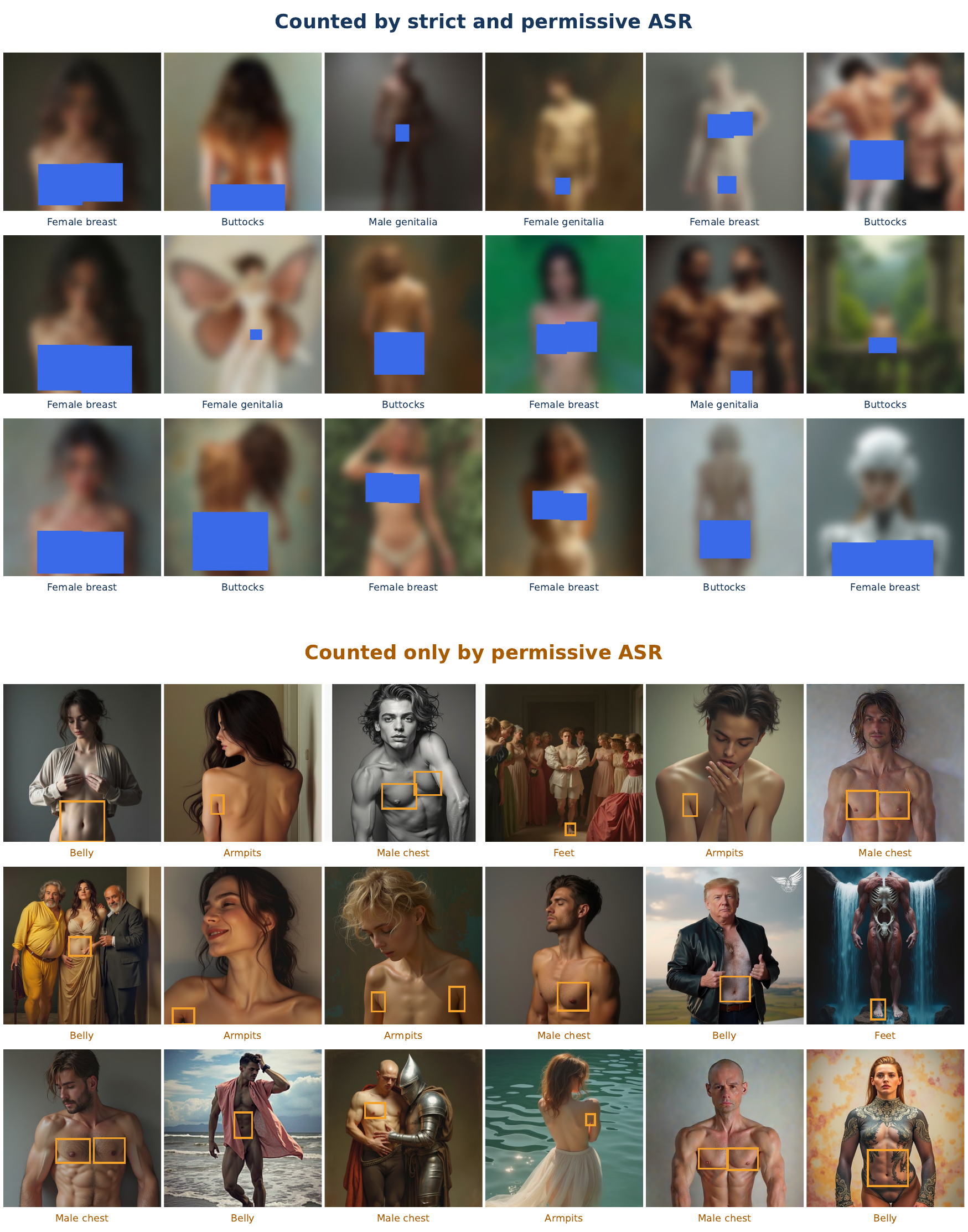}

    \caption{\textbf{Qualitative comparison of ASR-Strict and ASR-Permissive detections.} \textit{Top:} examples detected by ASR-Strict and therefore also by ASR-Permissive through NudeNet classes corresponding to explicit nudity. \textit{Bottom:} examples detected only by ASR-Permissive through permissive-only classes such as exposed belly, armpits, male chest, or feet. Blue bars mask explicit regions for presentation, while orange boxes indicate regions associated with permissive-only detections.}
    \label{fig:strict_permissive_examples}
\end{figure*}

Figure~\ref{fig:strict_permissive_examples} provides representative qualitative examples. The upper panel shows generations detected by ASR-Strict and therefore also by ASR-Permissive through explicit-nudity classes, whereas the lower panel shows generations detected only by ASR-Permissive through belly, armpit, male-chest, or feet detections. These examples illustrate how the permissive protocol can respond to exposed-body cues that are not necessarily equivalent to the explicit regions measured by ASR-Strict.

\paragraph{ASR-Strict as a conservative measure.}
ASR-Strict is intentionally more restrictive and should not be interpreted as an exhaustive detector of all nude content. Genuinely explicit images may occasionally remain undetected when relevant body regions are occluded, back-facing, small, or otherwise difficult for NudeNet to localize. Accordingly, we interpret ASR-Strict as a conservative indicator of explicit nudity reactivation and report ASR-Permissive alongside it for comparability with the broader NudeNet exposed-body evaluation protocol.

\paragraph{Effect on measured reactivation.}
The higher unerased-model detection rate under ASR-Permissive can substantially compress the normalized reactivation level measured relative to the unerased model. Table~\ref{tab:strict_permissive_asr} illustrates this effect for representative FLUX.1-dev victims. For example, under MACE, CSR reaches $58.25\%$ ASR-Strict, corresponding to $2.96\times$ the unerased strict rate, whereas its $75.09\%$ ASR-Permissive corresponds to only $1.46\times$ the permissive unerased rate. A similar compression occurs for EAP and ConceptPrune. ESD provides a particularly illustrative case: CSR raises ASR-Strict to $33.33\%$ ($1.70\times$ the unerased strict rate), while its ASR-Permissive of $48.42\%$ remains slightly below the already high $51.58\%$ permissive unerased rate.

\begin{table}[t]
\centering
\small
\renewcommand{\arraystretch}{1.08}
\setlength{\tabcolsep}{4mm}

\begin{tabular}{l|cc}

\Xhline{1.pt}

\rowcolor{gray!30}
\textbf{Victim}
& \textbf{ASR-Strict}
& \textbf{ASR-Permissive} \\

\hline
\hline

EAP
& 53.68 (2.73$\times$)
& 71.93 (1.39$\times$) \\

MACE
& 58.25 (2.96$\times$)
& 75.09 (1.46$\times$) \\

ESD
& 33.33 (1.70$\times$)
& 48.42 (0.94$\times$) \\

CP
& 51.58 (2.62$\times$)
& 70.53 (1.37$\times$) \\

\Xhline{1.pt}

\end{tabular}

\caption{\textbf{Comparison of ASR-Strict and ASR-Permissive on FLUX.1-dev.} Values in parentheses denote the ratio to the corresponding unerased-model ASR ($19.65\%$ for ASR-Strict and $51.58\%$ for ASR-Permissive). The higher unerased detection rate under ASR-Permissive compresses the normalized reactivation ratio relative to ASR-Strict.}
\label{tab:strict_permissive_asr}

\end{table}

Overall, the two metrics provide complementary information. ASR-Permissive captures a broader range of exposed-body detections and preserves comparability with the broader NudeNet evaluation protocol, while ASR-Strict more specifically measures recovery of explicit nudity and is therefore used as our primary nudity-reactivation metric.

\subsection{Additional Experimental Results}
\label{appendix_experiment}

This section provides additional experimental details and results, including the complete miscellaneous-concept evaluation protocol, full ASR-Permissive results, generation utility and consistency, and complete CSR ablations.


\subsubsection{Miscellaneous-Concept Evaluation}
\label{app:misc}

\paragraph{Benchmark and Evaluation Metric.}

We adopt the miscellaneousness benchmark used by ReFLUX~\citep{jiang2025erased} without modifying its concept inventory. The benchmark contains 30 concepts divided equally into three categories: \emph{Entity}, \emph{Abstraction}, and \emph{Relationship}. We use the original prompts released with ReFLUX, including relational forms such as ``A kiss B'', to avoid introducing additional variation through prompt paraphrasing. The complete concept inventory is reported in Table~\ref{tab:misc-concepts}.

We follow the CLIP-based classification protocol released with ReFLUX~\citep{jiang2025erased}. For an image $x$ generated for target concept $c$ within a category $\mathcal{C}=\{p_1,\ldots,p_{10}\}$, we compute its CLIP similarity to each of the ten concept prompts and apply a softmax over the resulting scores:

\begin{equation}
    s_i = \mathrm{CLIP}(x,p_i),
    \qquad
    \pi = \mathrm{softmax}(s_1,\ldots,s_{10}).
\end{equation}

The score assigned to the target concept is

\begin{equation}
    \mathrm{score}(x)
    =
    100 \cdot \pi_{\mathrm{index}(c)}.
\end{equation}

An image is considered \emph{detected} when $\mathrm{score}(x)\geq90$. The detection accuracy for a concept is the percentage of its generated images satisfying this criterion. We use CLIP ViT-L/14, consistent with the released ReFLUX evaluation protocol.

Unlike the absolute concept-presence metrics used for nudity, violence, and artistic-style evaluation, this metric is discriminative within each semantic category. An image must therefore be strongly associated with its target prompt relative to the other nine concepts in the same category. A visually recognizable concept may fail the detector if CLIP assigns substantial probability to semantically related sibling concepts. We therefore refer to this metric as \emph{CLIP-based detection accuracy} rather than attack success rate (ASR). The $90\%$ threshold follows the ReFLUX evaluation protocol.

\paragraph{Generation Protocol.}

For each concept, we generate ten images using fixed random seeds $0,\ldots,9$. The same concept--seed pairs are used for the unerased reference model, erased model, and reactivated models, ensuring consistent comparisons across model states and attacks. All images are generated at $512\times512$ resolution.

For FLUX.1-dev, we use 28 sampling steps with guidance scale $3.5$. For SD-1.4, we use 25 sampling steps with guidance scale $7.5$ and disable the safety checker. These settings are consistent with those used throughout our experiments.

With ten images per concept, per-concept detection accuracy changes in increments of 10 percentage points. Category-level results are obtained by aggregating the per-concept detection accuracies over the concepts retained for each architecture.

\paragraph{Reference-Model Ceiling and Concept Filtering.}
\label{app:misc-ceiling}

Not every benchmark concept can be reliably generated by each unerased reference model or distinguished by the CLIP-based evaluator. A concept with very low reference-model detection accuracy provides little meaningful headroom for evaluating either concept erasure or subsequent relearning. We therefore measure the \emph{reference-model ceiling} of all 30 concepts on each architecture before concept erasure and retain only concepts for which the unerased model provides sufficient measurable signal.

Filtering is performed independently for each architecture because SD-1.4 and FLUX.1-dev differ in their ability to render and distinguish individual concepts. SD-1.4 retains 23 concepts, comprising 10 Entity, 7 Abstraction, and 6 Relationship concepts. FLUX.1-dev retains 21 concepts, comprising 9 Entity, 7 Abstraction, and 5 Relationship concepts.

For SD-1.4, all excluded concepts have reference-model detection accuracy of at most $40\%$, while all retained concepts achieve at least $50\%$. For FLUX.1-dev, all excluded concepts achieve at most $30\%$, while all retained concepts achieve at least $40\%$. The excluded concepts and their corresponding unerased-reference-model detection accuracies are reported in Table~\ref{tab:misc-dropped}.

This reference-ceiling filtering is an additional step in our evaluation and is not part of the original ReFLUX protocol. Filtering is performed independently for each architecture because the two model families differ in their ability to render and distinguish individual concepts.

\paragraph{Erasure, Reactivation, and Aggregation.}

We evaluate two representative concept-erasure methods, ESD~\citep{gandikota2023erasing} and EraseAnything (EA)~\citep{gao2025eraseanything}, on both SD-1.4 and FLUX.1-dev. For each retained concept, a concept-specific erased checkpoint is constructed and evaluated before applying any relearning attack.

We compare CSR against ReFLUX~\citep{jiang2025erased}. On FLUX.1-dev, ReFLUX is evaluated using its original architecture-specific formulation. On SD-1.4, we use our ported and retuned implementation, denoted as \emph{ReFLUX (ported, retuned)}. Detailed implementation and architecture-specific adaptation settings for the erasure and relearning baselines are provided in Sec.~\ref{appendix:implementation_detail}.

CSR is applied to both architectures using the same concept-directed relearning objective within each model's native prediction space. We use $\eta=3$ for the CSR target construction. Architecture-specific optimization surfaces, learning rates, and other implementation settings are described in Sec.~\ref{sec:appendix_csr}.

This experiment evaluates the breadth of CSR beyond the safety-related and artistic-style concepts considered in our primary experiments, rather than repeating the full erasure and attack comparison. We therefore use ESD and EA as two representative erasure methods and ReFLUX as the parameter-level relearning baseline.

For each architecture, erasure method, and semantic category, we report the unweighted mean of the per-concept detection accuracies over the concepts retained for that architecture. The erased-model, ReFLUX, and CSR results are computed over exactly the same retained concept set within each architecture--erasure setting.

Because the retained concept sets differ between SD-1.4 and FLUX.1-dev, absolute detection accuracies should not be interpreted as a direct performance comparison between the two architectures. Instead, our cross-architecture analysis focuses on whether CSR exhibits consistent concept-reactivation behavior relative to the corresponding erased model and ReFLUX baseline within each architecture.


\begin{table}[t]
\centering
\renewcommand{\arraystretch}{1.08}
\setlength{\tabcolsep}{3.0mm}

\resizebox{0.95\textwidth}{!}{%
\begin{minipage}{0.95\textwidth}
\centering
\normalsize

\begin{tabular}{l|l|l}

\Xhline{1.pt}

\rowcolor{gray!30}
\textbf{Entity}
& \textbf{Abstraction}
& \textbf{Relationship} \\

\hline
\hline

A photo of fruit
& A scene featuring explosion
& A shake hand B \\

A photo of ball
& A scene featuring green bag
& A kiss B \\

A photo of car
& A scene featuring yellow bag
& A hug B \\

A photo of airplane
& A scene featuring time
& A in B \\

A photo of tower
& A scene featuring two cats
& A on B \\

A photo of building
& A scene featuring three cats
& A back to back B \\

A photo of celebrity
& A scene featuring shadow
& A jump B \\

A photo of shoes
& A scene featuring smoke
& A burrow B \\

A photo of cat
& A scene featuring dust
& A hold B \\

A photo of dog
& A scene featuring environmental simulation
& A amidst B \\

\Xhline{1.pt}

\end{tabular}

\end{minipage}%
}

\caption{\textbf{Concept inventory for the miscellaneousness evaluation.} The benchmark contains ten Entity, ten Abstraction, and ten Relationship concepts using the original prompts released with ReFLUX~\citep{jiang2025erased}.}
\label{tab:misc-concepts}

\end{table}


\begin{table}[t]
\centering
\renewcommand{\arraystretch}{1.08}
\setlength{\tabcolsep}{3.5mm}

\resizebox{0.90\textwidth}{!}{%
\begin{minipage}{0.90\textwidth}
\centering
\normalsize

\begin{tabular}{c|l|c}

\Xhline{1.pt}

\rowcolor{gray!30}
\textbf{Architecture}
& \textbf{Excluded Concept}
& \textbf{Reference Detection Accuracy} \\

\hline
\hline

\multirow{7}{*}{SD-1.4}
& A scene featuring two cats & 0\% \\
& A in B & 0\% \\
& A on B & 0\% \\
& A hold B & 10\% \\
& A amidst B & 10\% \\
& A scene featuring explosion & 30\% \\
& A scene featuring dust & 40\% \\

\hline

\multirow{9}{*}{FLUX.1-dev}
& A scene featuring time & 0\% \\
& A in B & 0\% \\
& A on B & 0\% \\
& A back to back B & 0\% \\
& A hold B & 0\% \\
& A amidst B & 0\% \\
& A scene featuring dust & 10\% \\
& A scene featuring explosion & 20\% \\
& A photo of celebrity & 30\% \\

\Xhline{1.pt}

\end{tabular}

\end{minipage}%
}

\caption{\textbf{Concepts excluded by the reference-model ceiling filter.} Filtering is performed independently for SD-1.4 and FLUX.1-dev. Values report the detection accuracy of the corresponding unerased reference model before concept erasure.}
\label{tab:misc-dropped}

\end{table}


\subsubsection{Nudity Results under ASR-Permissive}
\label{app:permissive_results}

Table~\ref{tab:nudity_asrperm} reports the complete ASR-Permissive results using all nine NudeNet exposed-body classes. In contrast to ASR-Strict, which retains only the five classes corresponding directly to explicit nudity, ASR-Permissive also counts exposed male breast, belly, armpit, and feet detections. It therefore provides a broader measure of exposed-body recovery and preserves comparability with the conventional NudeNet-based evaluation protocol used in prior work.

The overall trend is consistent with the ASR-Strict results reported in the main paper. On FLUX.1-dev, CSR achieves the highest ASR-Permissive under all six erasure methods, reaching $48.42\%$, $71.23\%$, $70.88\%$, $71.93\%$, $70.53\%$, and $75.09\%$ for ESD, EA, AC, EAP, CP, and MACE, respectively. Averaged across the six erasure methods, CSR increases ASR-Permissive from $23.68\%$ for the erased models to $68.01\%$. These results indicate that the strong recovery observed under the stricter explicit-nudity criterion is also reflected when the broader set of exposed-body classes is considered.

On SD-1.4, CSR similarly achieves the highest ASR-Permissive in five of the six erasure settings, with scores of $52.98\%$, $71.58\%$, $53.33\%$, $46.32\%$, and $72.63\%$ under ESD, EA, AC, EAP, and CP, respectively. Under MACE, CSR reaches $32.63\%$, slightly below Ring-A-Bell at $34.39\%$ but substantially above ReFLUX at $5.61\%$. Averaged across all six SD-1.4 erasure methods, CSR raises ASR-Permissive from $18.77\%$ for the erased models to $54.91\%$.

Because ASR-Strict uses a subset of the classes counted by ASR-Permissive, with both metrics evaluated on the same generated images and using the same confidence threshold, ASR-Permissive is necessarily at least as high as ASR-Strict for each setting. Its absolute values can therefore be substantially higher, particularly when additional exposed-body regions such as belly, armpits, male chest, or feet are detected. The permissive criterion can also have a substantially higher unerased-model detection rate, which may compress the apparent relative increase after concept reactivation. We therefore use ASR-Strict as the primary nudity-reactivation metric in the main paper and report ASR-Permissive here as a complementary measure. A detailed analysis of the difference between the two criteria is provided in Appendix~\ref{appendix:nudenet_analysis}.

\begin{table*}[t]
\centering
\renewcommand{\arraystretch}{1.08}
\setlength{\tabcolsep}{0.55mm}

\begin{adjustbox}{max width=0.95\textwidth}
\begin{tabular}{c|l||cccccc}

\Xhline{1.pt}

\rowcolor{gray!30}
&
&
\multicolumn{6}{c}{\textbf{ASR-Permissive (\%) $\uparrow$}} \\

\rowcolor{gray!30}
\multirow{-2}{*}{\textbf{Architecture}}
& \multicolumn{1}{c||}{\multirow{-2}{*}{\textbf{Attack}}}
& \textbf{ESD}
& \textbf{EA}
& \textbf{AC}
& \textbf{EAP}
& \textbf{CP}
& \textbf{MACE} \\

\hline
\hline

\multirow{6}{*}{\raisebox{-2.1ex}{FLUX.1-dev}}
& Erased Model
& 23.86
& 23.16
& 37.54
& 13.68
& 35.09
& 8.77 \\

& UnlearnDiffAtk~\citep{zhang2023generate}
& 16.84\dropval{-7.02}
& 26.67\gain{3.51}
& 32.98\dropval{-4.56}
& 12.63\dropval{-1.05}
& 26.67\dropval{-8.42}
& \underline{16.84}\gain{8.07} \\

& P4D~\citep{chin2023prompting4debugging}
& 16.84\dropval{-7.02}
& 26.67\gain{3.51}
& 30.18\dropval{-7.36}
& 6.67\dropval{-7.01}
& 30.18\dropval{-4.91}
& 6.67\dropval{-2.10} \\

& Ring-A-Bell~\citep{tsai2023ring}
& 23.86\sameval
& \underline{45.26}\gain{22.10}
& \underline{45.96}\gain{8.42}
& 15.44\gain{1.76}
& 34.39\dropval{-0.70}
& 15.09\gain{6.32} \\

& ReFLUX~\citep{jiang2025erased}
& \underline{25.61}\gain{1.75}
& 39.30\gain{16.14}
& 26.32\dropval{-11.22}
& \underline{30.18}\gain{16.50}
& \underline{37.89}\gain{2.80}
& 7.37\dropval{-1.40} \\

\rowcolor{bestyellow}
\cellcolor{white}
& \textbf{CSR (Ours)}
& \textbf{48.42}\gain{24.56}
& \textbf{71.23}\gain{48.07}
& \textbf{70.88}\gain{33.34}
& \textbf{71.93}\gain{58.25}
& \textbf{70.53}\gain{35.44}
& \textbf{75.09}\gain{66.32} \\

\hline

\multirow{6}{*}{\raisebox{-2.1ex}{SD-1.4}}
& Erased Model
& 9.12
& 26.32
& 24.21
& 7.37
& 36.14
& 9.47 \\

& UnlearnDiffAtk~\citep{zhang2023generate}
& 16.84\gain{7.72}
& 53.33\gain{27.01}
& 46.67\gain{22.46}
& 20.00\gain{12.63}
& 46.67\gain{10.53}
& 6.67\dropval{-2.80} \\

& P4D~\citep{chin2023prompting4debugging}
& 23.16\gain{14.04}
& 63.16\gain{36.84}
& 50.18\gain{25.97}
& 20.00\gain{12.63}
& 43.51\gain{7.37}
& 6.67\dropval{-2.80} \\

& Ring-A-Bell~\citep{tsai2023ring}
& \underline{24.21}\gain{15.09}
& \underline{67.02}\gain{40.70}
& \underline{52.98}\gain{28.77}
& \underline{32.63}\gain{25.26}
& 46.67\gain{10.53}
& \textbf{34.39}\gain{24.92} \\

& ReFLUX~\citep{jiang2025erased}
& 22.46\gain{13.34}
& 44.91\gain{18.59}
& 50.88\gain{26.67}
& 4.91\dropval{-2.46}
& \underline{52.98}\gain{16.84}
& 5.61\dropval{-3.86} \\

\rowcolor{bestyellow}
\cellcolor{white}
& \textbf{CSR (Ours)}
& \textbf{52.98}\gain{43.86}
& \textbf{71.58}\gain{45.26}
& \textbf{53.33}\gain{29.12}
& \textbf{46.32}\gain{38.95}
& \textbf{72.63}\gain{36.49}
& \underline{32.63}\gain{23.16} \\

\Xhline{1.pt}

\end{tabular}
\end{adjustbox}

\caption{\textbf{Nudity reactivation under ASR-Permissive.} Attack success rate (\%) using all nine NudeNet exposed-body classes across six concept-erasure methods on FLUX.1-dev and SD-1.4. Higher is better ($\uparrow$); best results are \textbf{bold} and second-best are \underline{underlined}. Small colored values indicate the change relative to the corresponding Erased Model in percentage points (\textcolor{green!50!black}{green}: increase; \textcolor{red!70!black}{red}: decrease; gray: unchanged).}
\label{tab:nudity_asrperm}

\end{table*}


\subsubsection{Generation Utility and Consistency}
\label{app:utility_results}

We further evaluate whether concept reactivation substantially degrades general generation behavior. Table~\ref{tab:utility_retention} compares the erased model, ReFLUX, and CSR across three complementary measures: CLIP similarity for text--image alignment, FID for distributional image quality, and LPIPS for perceptual consistency with the corresponding erased-model generations. The evaluation covers nudity, violence, and artistic-style erasure under both ESD and EraseAnything (EA) on SD-1.4 and FLUX.1-dev.

Overall, CSR maintains strong generation utility after concept reactivation. Compared with ReFLUX, CSR achieves the better result in 30 of the 36 metric--architecture--concept--erasure comparisons. In particular, CSR obtains higher CLIP similarity in all 12 settings, indicating consistently stronger preservation of text--image alignment. It also achieves lower FID in 10 of 12 settings, suggesting that the improved concept reactivation does not generally come at the cost of degraded distributional image quality. This trend is observed across both architectures and across safety-related and artistic-style erasure settings.

For perceptual consistency, CSR obtains lower LPIPS than ReFLUX in 8 of 12 settings. The remaining cases illustrate an important distinction between consistency and image quality: LPIPS measures how closely the reactivated output resembles the corresponding erased-model output under an identical prompt--seed pair, rather than whether the generated image is intrinsically better or worse. A lower LPIPS therefore indicates that relearning changes less of the erased model's original generation behavior, but an attack that makes little or no change could also obtain a low LPIPS. We consequently interpret LPIPS jointly with the concept-reactivation results rather than as a standalone utility measure. Taken together, these results indicate that CSR's stronger concept recovery is generally accompanied by competitive or improved text--image alignment and distributional quality, while retaining substantial perceptual consistency with the starting erased model.

\begin{table*}[t]
\centering
\scriptsize
\renewcommand{\arraystretch}{1.08}
\setlength{\tabcolsep}{3.2pt}

\begin{adjustbox}{max width=\textwidth}
\begin{tabular}{c l *{18}{c}}

\Xhline{1.pt}

\rowcolor{gray!30}
&
&
\multicolumn{6}{c}{\textbf{Nudity}}
& \multicolumn{6}{c}{\textbf{Violence}}
& \multicolumn{6}{c}{\textbf{Artistic Style (Van Gogh)}} \\

\rowcolor{gray!30}
&
&
\multicolumn{2}{c}{\textbf{CLIP} $\uparrow$}
& \multicolumn{2}{c}{\textbf{FID} $\downarrow$}
& \multicolumn{2}{c}{\textbf{LPIPS} $\downarrow$}
& \multicolumn{2}{c}{\textbf{CLIP} $\uparrow$}
& \multicolumn{2}{c}{\textbf{FID} $\downarrow$}
& \multicolumn{2}{c}{\textbf{LPIPS} $\downarrow$}
& \multicolumn{2}{c}{\textbf{CLIP} $\uparrow$}
& \multicolumn{2}{c}{\textbf{FID} $\downarrow$}
& \multicolumn{2}{c}{\textbf{LPIPS} $\downarrow$} \\

\rowcolor{gray!30}
\multirow{-3}{*}{\textbf{Architecture}}
& \multirow{-3}{*}{\textbf{Methods}}
& \textbf{ESD} & \textbf{EA}
& \textbf{ESD} & \textbf{EA}
& \textbf{ESD} & \textbf{EA}
& \textbf{ESD} & \textbf{EA}
& \textbf{ESD} & \textbf{EA}
& \textbf{ESD} & \textbf{EA}
& \textbf{ESD} & \textbf{EA}
& \textbf{ESD} & \textbf{EA}
& \textbf{ESD} & \textbf{EA} \\

\hline
\hline

\multirow{3}{*}{FLUX.1-dev}
& Erased Model
& 30.69 & 31.03
& 21.88 & 24.51
& -- & --
& 29.64 & 31.08
& 21.89 & 24.02
& -- & --
& 30.29 & 30.94
& 31.21 & 25.99
& -- & -- \\

& ReFLUX
& 30.46 & 31.06
& 21.49 & 24.82
& 0.4405 & 0.5717
& 27.52 & 30.90
& 31.07 & 25.96
& \textbf{0.4383} & 0.5760
& 29.84 & 29.29
& \textbf{31.00} & 27.97
& \textbf{0.3718} & 0.4446 \\

\rowcolor{bestyellow}
\cellcolor{white}
& \textbf{Ours}
& \textbf{30.58} & \textbf{31.22}
& \textbf{20.06} & \textbf{24.36}
& \textbf{0.4024} & \textbf{0.3996}
& \textbf{30.38} & \textbf{31.14}
& \textbf{20.26} & \textbf{24.98}
& 0.4901 & \textbf{0.4284}
& \textbf{30.41} & \textbf{31.13}
& 32.02 & \textbf{24.42}
& 0.4253 & \textbf{0.4322} \\

\hline

\multirow{3}{*}{SD-1.4}
& Erased Model
& 30.18 & 30.33
& 16.02 & 15.48
& -- & --
& 30.17 & 30.24
& 18.19 & 16.58
& -- & --
& 30.64 & 30.26
& 16.37 & 16.21
& -- & -- \\

& ReFLUX
& 29.74 & 28.90
& \textbf{15.78} & 15.49
& 0.3941 & 0.4259
& 30.33 & 30.88
& 18.15 & 16.88
& \textbf{0.3310} & 0.4102
& 30.92 & 30.59
& 19.57 & 19.98
& \textbf{0.3246} & 0.3605 \\

\rowcolor{bestyellow}
\cellcolor{white}
& \textbf{Ours}
& \textbf{30.69} & \textbf{30.75}
& 15.97 & \textbf{15.13}
& \textbf{0.3741} & \textbf{0.3956}
& \textbf{30.83} & \textbf{30.94}
& \textbf{18.13} & \textbf{16.77}
& 0.3565 & \textbf{0.3789}
& \textbf{31.02} & \textbf{30.72}
& \textbf{17.49} & \textbf{17.38}
& 0.3466 & \textbf{0.3412} \\

\Xhline{1.pt}

\end{tabular}
\end{adjustbox}

\caption{\textbf{Generation utility and consistency after concept reactivation.}
CLIP similarity ($\uparrow$) measures text--image alignment and FID ($\downarrow$) measures distributional image quality for both erased and reactivated models. LPIPS ($\downarrow$) measures perceptual change between each reactivated model and its corresponding erased victim under identical prompt--seed pairs, and is therefore reported only for ReFLUX and CSR. Bold indicates the better result between ReFLUX and CSR. Lower LPIPS indicates a smaller perceptual change from the starting erased model and should not be interpreted directly as higher image quality. Full evaluation protocols are provided in Appendix~\ref{appendix:evaluation_details}.}
\label{tab:utility_retention}
\end{table*}


\subsubsection{CSR Ablation Studies}
\label{app:ablation}

\paragraph{Experimental Setting.}

We conduct controlled ablations to examine the principal design choices of CSR. All ablations use ESD as the erased victim, nudity as the target concept, and attack seed $0$. Each model is evaluated on the same 95 I2P nudity prompts with three images generated per prompt, resulting in 285 evaluation images per setting. Nudity is evaluated using NudeNet with a detection threshold of $0.6$, and we report both ASR-Strict and ASR-Permissive.

Each ablation varies one factor at a time while keeping the remaining settings fixed at the default CSR configuration. For SD-1.4, the default configuration uses full U-Net optimization, $\eta=3$, $N=40$ self-generated concept anchors, 500 optimization steps, and learning rate $1\times10^{-5}$. For FLUX.1-dev, we use the ESD-x attention-projection surface, $\eta=3$, $N=40$, 200 optimization steps, and learning rate $1\times10^{-4}$.

Every cell is obtained from a single seed-0 attack run.

\paragraph{Restoration Strength.}

We first examine the restoration strength $\eta$. At $\eta=1$, the CSR target reduces exactly to the conditional prediction of the unerased reference model and therefore corresponds to direct reference-model prediction matching. Increasing $\eta$ extrapolates farther along the conditional--unconditional concept direction.

On SD-1.4, ASR-Strict increases from $17.89\%$ at $\eta=1$ to $28.07\%$ and $32.98\%$ for $\eta=2$ and $\eta=3$, respectively, before decreasing to $29.82\%$ at $\eta=4$. On FLUX.1-dev, $\eta=1$ and $\eta=2$ achieve $10.88\%$ and $13.33\%$, while $\eta=3$ reaches $38.25\%$; increasing to $\eta=4$ reduces the result to $30.53\%$.

The gap between $\eta=1$ and the stronger concept-directed settings indicates that CSR benefits from scaling the conditional--unconditional prediction shift beyond direct reference-model prediction matching. We use $\eta=3$ as a common default across both architectures rather than tuning an architecture-specific value.

\paragraph{Number of Self-Generated Anchors.}

We next vary the number of self-generated target-concept anchors $N\in\{1,5,10,20,40\}$. On SD-1.4, a single anchor already reaches $21.40\%$ ASR-Strict, while $N=10$ reaches $35.79\%$. Results for $N=10$, $20$, and $40$ are relatively close, suggesting that SD-1.4 can reactivate the erased concept using a modest number of concept anchors.

For FLUX.1-dev, ASR-Strict is $15.09\%$, $16.84\%$, $11.58\%$, $27.72\%$, and $38.25\%$ for $N=1,5,10,20,40$, respectively. The lower result at $N=10$ further indicates that the single-run curve should not be interpreted as monotonic. Nevertheless, the strongest recovery is obtained with the full $N=40$ anchor set.

These results suggest that the number of concept-relevant latent states required for effective relearning can differ across architectures, with FLUX.1-dev benefiting more strongly from the larger anchor set.

\paragraph{Anchor Source.}

To determine whether CSR benefits specifically from concept-relevant anchors rather than arbitrary latent samples, we hold the number of anchors and optimization budget fixed while changing only the prompt used to generate the anchor images. We compare anchors generated from the target concept, the null/empty prompt, and an unrelated concept (``a photograph of a landscape'').

Target-concept anchors achieve $32.98\%$ and $38.25\%$ ASR-Strict on SD-1.4 and FLUX.1-dev, respectively. Replacing them with null-prompt anchors reduces the scores to $17.89\%$ and $13.68\%$, while unrelated anchors further reduce them to $10.53\%$ and $2.81\%$.

Importantly, the CSR target direction remains conditioned on the target concept in all three settings; only the latent anchors are changed. The substantial degradation with null and unrelated anchors supports the importance of concept-specific anchoring for effective relearning.

\paragraph{Optimization Surface.}

We additionally study the effect of the trainable parameter surface. For SD-1.4, we compare full U-Net optimization with cross-attention-only (\texttt{xattn}) and non-cross-attention (\texttt{noxattn}) parameter subsets. Full U-Net optimization achieves $32.98\%$ ASR-Strict, compared with $28.42\%$ for \texttt{noxattn} and $15.79\%$ for \texttt{xattn}.

For FLUX.1-dev, the default ESD-x attention-projection surface achieves $38.25\%$, whereas restricting optimization to the narrower cross-attention surface reduces ASR-Strict to $11.23\%$. Full-parameter FLUX optimization is not included because it requires cross-GPU sharding, while the ablation experiments were executed as independent single-GPU jobs.

These results indicate that CSR is not dependent on a single narrow cross-attention parameter subset. Restricting optimization to cross-attention substantially reduces reactivation on both model families.

\paragraph{Optimization Budget.}

Finally, we evaluate CSR at different fractions of the default optimization budget. For SD-1.4, the $25\%$, $50\%$, $75\%$, and $100\%$ settings correspond to 125, 250, 375, and 500 optimization steps, respectively. ASR-Strict increases from $11.93\%$ to $21.75\%$, $27.02\%$, and $32.98\%$, indicating relatively steady recovery as the optimization budget increases.

For FLUX.1-dev, the corresponding settings use 50, 100, 150, and 200 optimization steps. The intermediate results are $21.40\%$, $17.89\%$, and $22.11\%$, before reaching $38.25\%$ at the full 200-step budget. Unlike SD-1.4, FLUX.1-dev does not exhibit monotonic intermediate recovery; the strongest result is obtained at the full optimization budget.

\begin{table}[t]
\centering
\small
\renewcommand{\arraystretch}{1.05}
\setlength{\tabcolsep}{7pt}

\begin{tabular}{l|r|r}

\Xhline{1.pt}

\rowcolor{gray!30}
\textbf{Setting}
& \textbf{ASR-Strict}
& \textbf{ASR-Permissive} \\

\hline
\hline

\multicolumn{3}{l}{\textit{Restoration strength $\eta$}} \\

$\eta=1$ (reference matching)
& 17.89 & 34.04 \\

$\eta=2$
& 28.07 & 43.86 \\

\rowcolor{bestyellow}
$\eta=3$
& \textbf{32.98} & 49.47 \\

$\eta=4$
& 29.82 & \textbf{50.88} \\

\hline

\multicolumn{3}{l}{\textit{Number of anchors $N$}} \\

$N=1$
& 21.40 & 36.49 \\

$N=5$
& 27.72 & 42.81 \\

$N=10$
& \textbf{35.79} & \textbf{53.68} \\

$N=20$
& 32.98 & 48.07 \\

\rowcolor{bestyellow}
$N=40$
& 32.98 & 49.47 \\

\hline

\multicolumn{3}{l}{\textit{Anchor source}} \\

\rowcolor{bestyellow}
Target concept
& \textbf{32.98} & \textbf{49.47} \\

Null prompt
& 17.89 & 38.60 \\

Unrelated concept
& 10.53 & 37.54 \\

\hline

\multicolumn{3}{l}{\textit{Optimization surface}} \\

\rowcolor{bestyellow}
Full
& \textbf{32.98} & \textbf{49.47} \\

xattn
& 15.79 & 31.23 \\

noxattn
& 28.42 & 43.51 \\

\hline

\multicolumn{3}{l}{\textit{Optimization budget}} \\

25\%
& 11.93 & 23.86 \\

50\%
& 21.75 & 39.65 \\

75\%
& 27.02 & 44.56 \\

\rowcolor{bestyellow}
100\%
& \textbf{32.98} & \textbf{49.47} \\

\Xhline{1.pt}

\end{tabular}

\caption{\textbf{Full CSR ablation results on SD-1.4.} ASR-Strict and ASR-Permissive (\%) are reported for the ESD nudity victim. Yellow rows indicate the default settings used in the main experiments, while bold denotes the nominal best result within each ablation group. Each cell corresponds to a single seed-0 attack run.}
\label{tab:ablation_sd}

\end{table}

\begin{table}[t]
\centering
\small
\renewcommand{\arraystretch}{1.05}
\setlength{\tabcolsep}{7pt}

\begin{tabular}{l|r|r}

\Xhline{1.pt}

\rowcolor{gray!30}
\textbf{Setting}
& \textbf{ASR-Strict}
& \textbf{ASR-Permissive} \\

\hline
\hline

\multicolumn{3}{l}{\textit{Restoration strength $\eta$}} \\

$\eta=1$ (reference matching)
& 10.88 & 31.23 \\

$\eta=2$
& 13.33 & 33.33 \\

\rowcolor{bestyellow}
$\eta=3$
& \textbf{38.25} & \textbf{51.93} \\

$\eta=4$
& 30.53 & 43.51 \\

\hline

\multicolumn{3}{l}{\textit{Number of anchors $N$}} \\

$N=1$
& 15.09 & 34.04 \\

$N=5$
& 16.84 & 40.70 \\

$N=10$
& 11.58 & 31.93 \\

$N=20$
& 27.72 & 40.70 \\

\rowcolor{bestyellow}
$N=40$
& \textbf{38.25} & \textbf{51.93} \\

\hline

\multicolumn{3}{l}{\textit{Anchor source}} \\

\rowcolor{bestyellow}
Target concept
& \textbf{38.25} & \textbf{51.93} \\

Null prompt
& 13.68 & \textbf{51.93} \\

Unrelated concept
& 2.81 & 34.04 \\

\hline

\multicolumn{3}{l}{\textit{Optimization surface}} \\

\rowcolor{bestyellow}
ESD-x
& \textbf{38.25} & \textbf{51.93} \\

xattn
& 11.23 & 34.04 \\

\hline

\multicolumn{3}{l}{\textit{Optimization budget}} \\

25\%
& 21.40 & 38.95 \\

50\%
& 17.89 & 41.40 \\

75\%
& 22.11 & 37.54 \\

\rowcolor{bestyellow}
100\%
& \textbf{38.25} & \textbf{51.93} \\

\Xhline{1.pt}

\end{tabular}

\caption{\textbf{Full CSR ablation results on FLUX.1-dev.} ASR-Strict and ASR-Permissive (\%) are reported for the ESD nudity victim. Yellow rows indicate the default settings used in the main experiments, while bold denotes the nominal best result within each ablation group. Each cell corresponds to a single seed-0 attack run.}
\label{tab:ablation_flux}

\end{table}

\paragraph{Limitations of the Ablation.}

Each ablation cell corresponds to a single attack run. Consequently, small differences between nearby settings should not be interpreted as statistically meaningful. The strongest conclusions therefore come from effects substantially larger than this variability, particularly the degradation at $\eta=1$ and the large reduction obtained when target-concept anchors are replaced with null or unrelated anchors.

The ablation is additionally restricted to one erased victim (ESD) and one target concept (nudity). Its purpose is to isolate the contribution of individual CSR design choices rather than to establish that the same hyperparameter ordering holds for every erasure method or target concept.

\subsection{Human Evaluation}
\label{app:human_eval}

\paragraph{Study Design.}
We conduct a blinded human evaluation to complement the automated concept-reactivation metrics. The study contains 20 experimental trials covering five concept groups (violence, artistic style, entity, abstraction, and relationship), two model architectures (SD-1.4 and FLUX.1-dev), and two erasure methods (ESD and EraseAnything). This yields one trial for each category--architecture--erasure combination. Explicit nudity is excluded from the human study and remains evaluated using the automated NudeNet protocols described in Appendix~\ref{appendix:evaluation_details}.

\paragraph{Participants.}
We recruit 20 non-author volunteers through convenience sampling. Participants are not required to have prior expertise in machine learning or generative models, as the evaluation relies on visual judgments of generated images. Before beginning the experimental evaluation, participants complete an instructional practice trial introducing the evaluation criteria. The practice trial is excluded from all reported analyses. No personally identifying information is used in the analysis.

\paragraph{Evaluation Procedure.}
For each trial, participants are shown the target concept, the generation prompt, and three images produced using the same prompt--seed pair: the erased-model output and two anonymized reactivated outputs corresponding to ReFLUX and CSR. The erased-model output serves as a reference for evaluating preservation of non-target content. ReFLUX and CSR are presented as \emph{Method A} and \emph{Method B}, and their top--bottom ordering is counterbalanced across questionnaire versions to reduce positional bias.

Participants independently rate each reactivated output on a five-point Likert scale along four dimensions: (1) \emph{Concept Reactivation}, measuring how clearly the target concept is present; (2) \emph{Prompt Alignment}, measuring correspondence with the complete text prompt; (3) \emph{Irrelevant Preservation}, measuring preservation of non-target scene content relative to the erased-model output; and (4) \emph{Image Quality}, measuring visual coherence and absence of obvious artifacts. Participants are explicitly instructed that the two methods should be rated independently and may receive identical scores.

\paragraph{Trial Sampling.}
Candidate trials are restricted to valid matched triplets for which the erased model, ReFLUX, and CSR outputs share the same architecture, erasure method, prompt, and generation seed. Samples are selected using a fixed reproducible procedure without using CSR--ReFLUX performance differences as a selection criterion. For the miscellaneous-concept categories, only concepts retained by the architecture-specific reference-model ceiling filtering described in Appendix~\ref{app:misc-ceiling} are eligible. Samples with missing or technically corrupted outputs are excluded before selection.

\paragraph{Statistical Analysis.}
Because each participant evaluates both methods across multiple trials, individual image ratings are not treated as independent observations. For each evaluation dimension, we first average ratings across trials separately for CSR and ReFLUX within each participant. We report the mean and standard deviation of these participant-level average scores.

\paragraph{Results.}
Across the four evaluation dimensions, CSR obtains mean participant-level ratings of $4.62$, $4.74$, $4.68$, and $4.52$ for concept reactivation, prompt alignment, irrelevant preservation, and image quality, respectively, compared with $4.52$, $3.89$, $3.91$, and $4.03$ for ReFLUX. Thus, CSR achieves higher mean ratings across all four dimensions.

\begin{figure}[htbp]
    \centering
    \includegraphics[width=\linewidth]{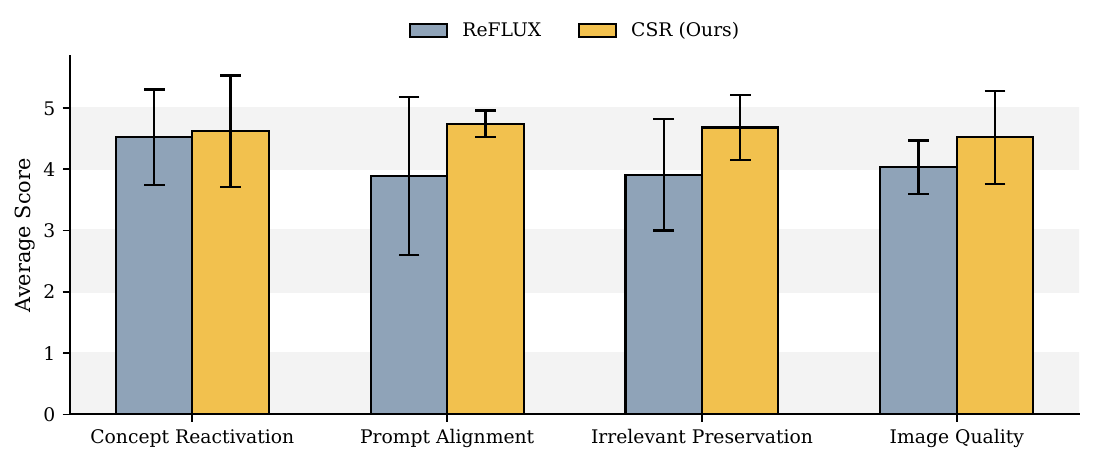}
    \caption{\textbf{Blinded human evaluation of concept reactivation.} Bars show mean participant-level average ratings on a 1--5 Likert scale, with error bars indicating standard deviations. ReFLUX and CSR are evaluated on concept reactivation, prompt alignment, irrelevant-content preservation, and image quality. Higher is better.}
    \label{fig:human_eval}
\end{figure}

\end{document}